\documentclass{article}

\usepackage{microtype}
\usepackage{graphicx}
\usepackage{subcaption}
\usepackage{booktabs} 

\usepackage{hyperref}

\usepackage[preprint]{icml2026}

\usepackage{mathtools}
\usepackage{xcolor}
\usepackage{colortbl}

\definecolor{customgray}{RGB}{240,240,240}
\usepackage{hyperref}

\usepackage{url}
\usepackage{graphicx}
\usepackage{amsthm}
\usepackage{amssymb}
\usepackage{amsmath}
\usepackage{multirow}
\usepackage{enumitem}
\usepackage{algorithm}

\usepackage{booktabs}
\usepackage{colortbl}
\usepackage{xcolor}
\usepackage{pifont}
\definecolor{rowblue}{RGB}{236,244,252}
\usepackage{tabularx}
\renewcommand{\algorithmiccomment}[1]{\hfill $\triangleright$ #1}

\usepackage[capitalize,noabbrev]{cleveref}

\theoremstyle{plain}
\newtheorem{theorem}{Theorem}[section]

\theoremstyle{definition}
\newtheorem{definition}[theorem]{Definition}

\theoremstyle{remark}

\usepackage[textsize=tiny]{todonotes}

\icmltitlerunning{Beyond Next-Token Alignment: Distilling Multimodal Large Language Models via Token Interactions}

\begin{document}
\twocolumn[
  \icmltitle{Beyond Next-Token Alignment: Distilling\\ Multimodal Large Language Models via Token Interactions}

  \begin{icmlauthorlist}
    \icmlauthor{Lin Chen}{cas,ucas,ant}
    \icmlauthor{Xiaoke Zhao}{ant}
    \icmlauthor{Kun Ding}{cas}
    \icmlauthor{Weiwei Feng}{ant,zju}
    \icmlauthor{Changtao Miao}{ant,zju}
    \icmlauthor{Zili Wang}{cas,ucas}\\
    \icmlauthor{Wenxuan Guo}{cas,ucas}
    \icmlauthor{Ying Wang}{cas}
    \icmlauthor{Kaiyuan Zheng}{ant}
    \icmlauthor{Bo Zhang}{ant}
    \icmlauthor{Zhe Li}{ant}
    \icmlauthor{Shiming Xiang}{cas,ucas}
  \end{icmlauthorlist}

  \icmlaffiliation{cas}{MAIS, Institute of Automation, Chinese Academy of Sciences}
  \icmlaffiliation{ucas}{School of Artificial Intelligence, University of Chinese Academy of Sciences}
  \icmlaffiliation{zju}{Zhejiang University}
  \icmlaffiliation{ant}{Ant Group}

  \icmlcorrespondingauthor{Kun Ding}{kun.ding@ia.ac.cn}
  \icmlcorrespondingauthor{Weiwei Feng}{fengww@mail.ustc.edu.cn}

  \icmlkeywords{Machine Learning, ICML}

  \vskip 0.3in
]

\footnotetext[1]{This work was done when Lin Chen was an intern at Ant Digital Technologies, Ant Group.}




\printAffiliationsAndNotice{}  

\begin{abstract}
Multimodal Large Language Models (MLLMs) demonstrate impressive cross-modal capabilities, yet their substantial size poses significant deployment challenges. Knowledge distillation (KD) is a promising solution for compressing these models, but existing methods primarily rely on static next-token alignment, neglecting the dynamic token interactions, which embed essential capabilities for multimodal understanding and generation. To this end, we introduce \textbf{Align-TI}, a novel KD framework designed from the perspective of \textbf{T}oken \textbf{I}nteractions. Our approach is motivated by the insight that MLLMs rely on two primary interactions: vision-instruction token interactions to extract relevant visual information, and intra-response token interactions for coherent generation. Accordingly, Align-TI introduces two components: IVA enables the student model to imitate the teacher's instruction-relevant visual information extract capability by aligning on salient visual regions. TPA captures the teacher's dynamic generative logic by aligning the sequential token-to-token transition probabilities. Extensive experiments demonstrate Align-TI's superiority. Notably, our approach achieves $2.6\%$ relative improvement over Vanilla KD, and our distilled Align-TI-2B even outperforms LLaVA-1.5-7B (a much larger MLLM) by $7.0\%$, establishing a new state-of-the-art distillation framework for training parameter-efficient MLLMs. Code is available at \href{https://github.com/lchen1019/Align-TI}{https://github.com/lchen1019/Align-TI}.
\end{abstract}

\begin{figure}[htbp]
    \centering
    \includegraphics[width=1\linewidth]{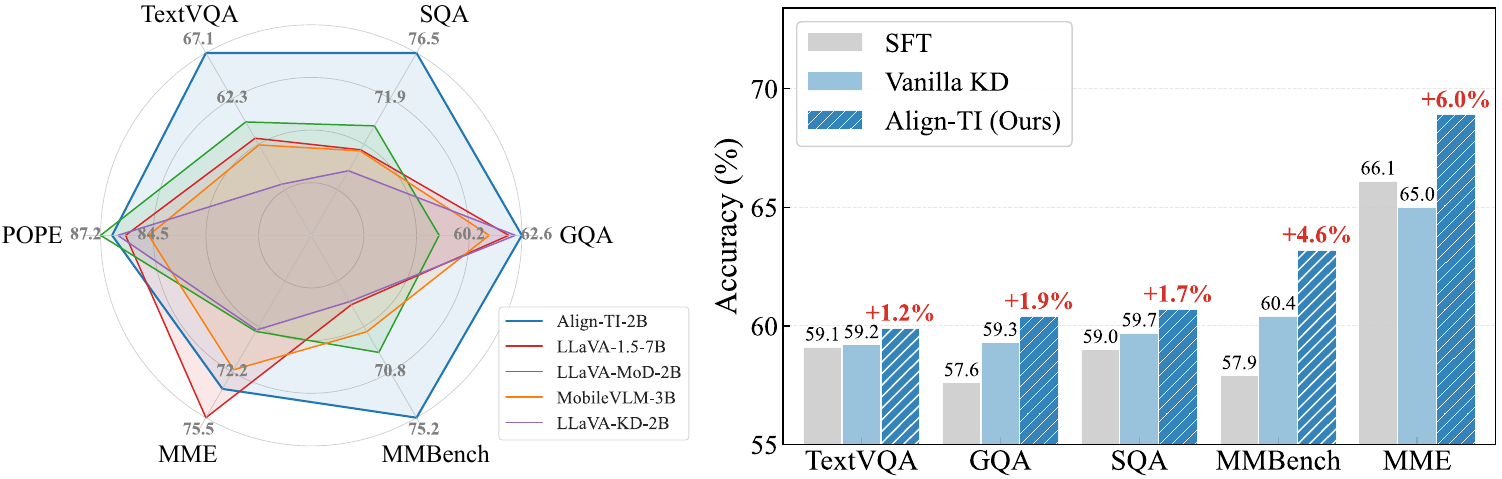}
    \caption{Experimental results overview. \textbf{Left:} Performance comparison between MLLMs distilled using our proposed Align-TI and other state-of-the-art MLLMs. \textbf{Right:} Performance gains achieved by Align-TI relative to the SFT and Vanilla KD baselines. (Details provided in Appendix~\ref{appendix:fig1-details}.)}
    \label{fig:intro-performance}
\end{figure}

\section{Introduction}

\begin{figure*}[t]
    \centering
    \includegraphics[width=0.96\linewidth]{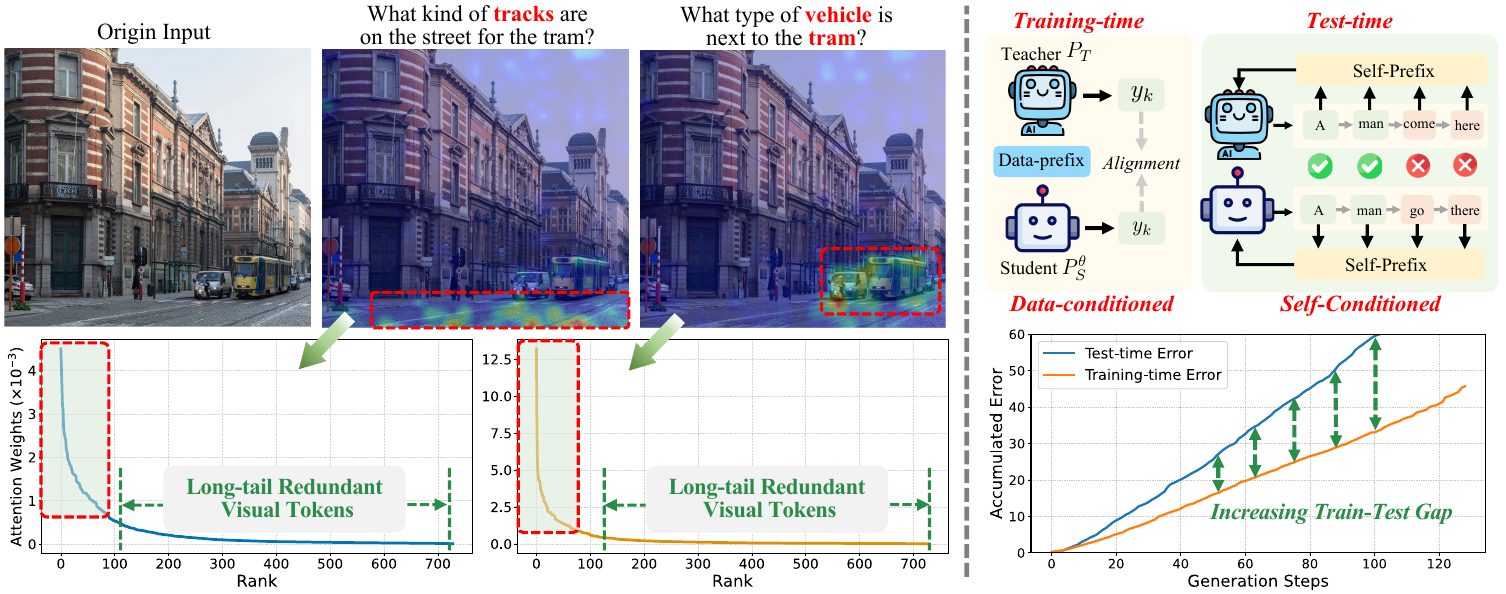}
    \vspace{-5pt}
    \caption{Motivation of MLLM distillation in view of token interactions. \textbf{Left: Vision-instruction token interaction analysis.} Visualizations of instruction-to-vision attention weights demonstrate that different instructions activate distinct visual focus areas, while exhibiting significant token redundancy. \textbf{Right: Intra-response token interaction analysis.} The discrepancy between data-conditioned prefix during training-time and self-conditioned prefix during test-time amplifies autoregressive accumulated error. (More details are provided in Appendix~\ref{sec:accerr-intro}.)}
    \vspace{-5pt}
    \label{fig:intro}
\end{figure*}

Multimodal Large Language Models (MLLMs)~\citep{liu2023visual, hurst2024gpt,guo2025seed1, comanici2025gemini} have emerged as a cornerstone in the pursuit of Artificial General Intelligence (AGI), showcasing remarkable capabilities in cross-modal understanding and generation. The success of contemporary MLLMs is predominantly built upon the foundation of autoregressive large language models~\citep{floridi2020gpt, touvron2023llama, yang2025qwen3}, which adopt self-supervised pretraining with the next-token prediction paradigm. Previous studies~\citep{radford2018improving, mann2020language, achiam2023gpt} have demonstrated that scaling up these models' parameters and training data continuously pushes the performance boundaries, while it also results in large-scale models with significant computational demands. Knowledge distillation (KD)~\citep{hinton2015distilling} provides a promising pathway to reduce computational overhead requirements by replicating large-scale teacher model capabilities in more parameter-efficient students through systematic knowledge transfer.

Prior work~\citep{xu2024llavadi} demonstrates that knowledge transfer in MLLM distillation via intermediate features or attention maps is often ineffective, primarily due to functional misalignment between student and teacher layers. In contrast, aligning output token distributions has proven to be a more effective way. Building upon the foundation of token-level alignment, subsequent research~\citep{cai2024llavakd, feng2025alignkd, shu2024llavamod} introduces additional components, such as MoE architectures~\citep{dai2024deepseekmoe}. \textbf{However, the token-level alignment in these methods remains limited to static next-token alignment, where the student only mimics the teacher's output distribution on fixed, off-policy sequences, neglecting to model the dynamic token interactions that encode critical capabilities for MLLM understanding and generation.} Specifically, these interactions encode crucial capabilities in two stages: (1) Prefilling: Vision-instruction token interactions encode instruction-aware visual information extraction capability. (2) Decoding: Intra-response token interactions encode dynamic reasoning and generation capability. The absence of such interaction modeling restricts the student to acquiring shallow statistical patterns from the teacher's outputs, rather than its deeper mechanisms for understanding and generation, thus resulting in insufficient knowledge transfer.

To provide better knowledge transfer, we further analyze the underlying characteristics of these two types of interactions. \textbf{(1) Vision-instruction token interactions.} In Fig.~\ref{fig:intro} (Left), we visualize the instruction-to-vision attention weights, it reveals two key observations: (a) Identical images elicit distinct region activations under different instructions. (b) Instruction tokens attend primarily to a few salient visual tokens, while the majority follow a long-tailed distribution. This imbalance indicates that prior distillation methods compel the capacity-constrained student to misallocate precious resources toward mimicking the teacher's processing of low-utility tokens, thereby hindering its ability to master instruction-critical representations. \textbf{(2) Intra-response token interactions.} As depicted in Fig.~\ref{fig:intro} (Right), prior distillation methods primarily align the teacher and student models' data-conditioned next token prediction probabilities, where the prefix originates from the static training corpus. This approach neglects the transition dynamics inherent in test-time generation, where predictions are conditioned on the self-generated outputs. As shown in Fig.~\ref{fig:intro} (Right), this issue leads to a widening accumulated error gap between training-time and test-time, a challenge referred to as exposure bias in imitation learning~\citep{ross2011reduction,arora2022exposure,kim2024promptkd}.

Based on the aforementioned discussion, we propose \textbf{Align-TI}, a framework that explicitly modeling KD for MLLMs from the perspective of token interactions. Align-TI consists of two core components: \textbf{I}nstruction-aware \textbf{V}ision \textbf{A}lignment (\textbf{IVA}) and \textbf{T}ransition \textbf{P}robability \textbf{A}lignment (\textbf{TPA}), corresponding to the two interaction types. Specifically, IVA enables the student model to learn on the teacher model’s instruction-aware visual focus, facilitating the transfer of the teacher's visual information extraction capability. Additionally, recognizing that this visual focus varies significantly across transformer layers, we design the \textbf{I}nstruction-\textbf{R}elevant \textbf{S}core (\textbf{IRS}) to quantify the relevance of a layer's attention map to the given instruction. This enables principled selection of the most instruction-relevant visual focus for IVA. Furthermore, TPA explicitly aligns the token-to-token transition probabilities, enabling the student to better learn the teacher's continuous token generation patterns. Moreover, experimental evidences demonstrate that TPA helps mitigate the teacher-student autoregressive generation discrepancy at test time.

\begin{figure*}[ht]
    \centering
    \includegraphics[width=0.94\linewidth]{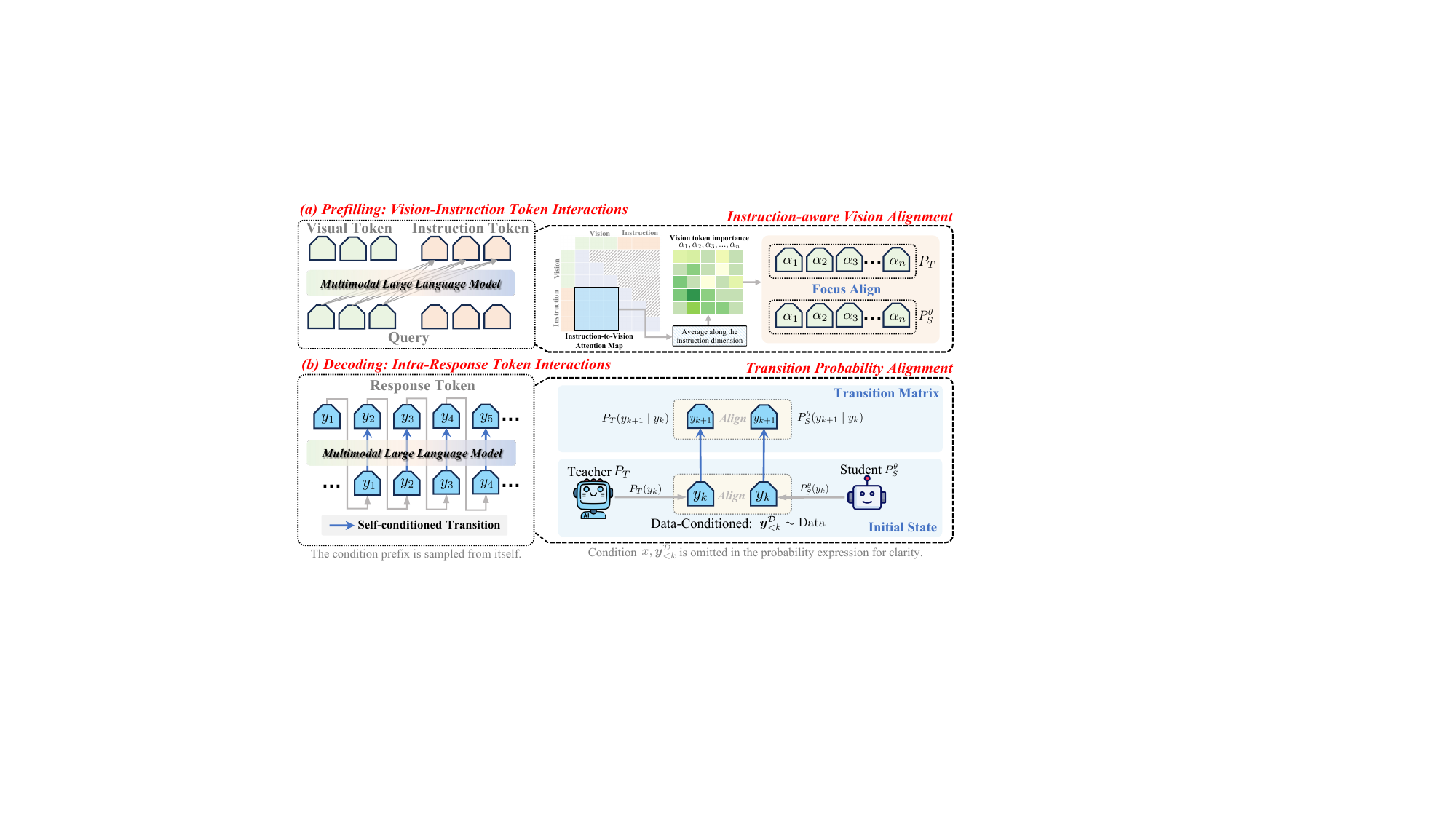}
    \vspace{-5pt}
    \caption{Overview of the proposed Align-TI. The framework explicitly models MLLM KD from the perspective of token interactions.}
    \label{fig:pipeline}
    \vspace{-5pt}
\end{figure*}

Extensive experiments validate the efficacy of Align-TI in distilling knowledge from large-scale MLLMs into compact models, as summarized in Fig.~\ref{fig:intro-performance}. Our distilled 1B model achieves relative improvements of $4.6\%$ on MMBench~\citep{liu2024mmbench} and $6.0\%$ on MME~\citep{fu2024mme} compared to the Vanilla KD baseline. Furthermore, our Align-TI-2B outperforms both LLaVA-MoD-2B~\citep{shu2024llavamod} (a strong MoE-based distillation baseline) and the larger LLaVA-1.5-7B~\citep{llava15}, advancing the state-of-the-art in distillation strategies for parameter-efficient MLLMs.

\section{Preliminaries}
\noindent \textbf{Multimodal Large Language Models.}
By integrating the capabilities of pretrained LLMs with vision encoders, MLLMs establish a connection between the visual and linguistic modalities. Given an input $\boldsymbol{x} = (X_v, X_q)$, where $X_v$ represents the input image and $X_q$ denotes the textual instruction, MLLMs aim to generate a response $\boldsymbol{y}$ conditioned on $\boldsymbol{x}$.  A typical MLLM architecture comprises three core components: a vision encoder, a vision-language projector, and a large language model. The MLLM processing pipeline can be formally defined as:
\begin{equation}
\mathcal{O} = \mathrm{LLM}_{\phi}\left( \mathrm{Proj}_{\omega} \left( \mathrm{Vis}_{\psi}(X_v) \right), X_q \right),
\end{equation}
where $\psi$, $\omega$ and $\phi$ denote the parameters of the vision encoder, projector and language model, respectively. The output $\mathcal{O} = \{\boldsymbol{v}, \boldsymbol{q}, \boldsymbol{y}\}$ consists of visual tokens $\boldsymbol{v}$, instruction tokens $\boldsymbol{q}$, and response tokens $\boldsymbol{y}$. Note that $\boldsymbol{v}$ and $\boldsymbol{q}$ are present in the output because the transformer decoder maintains an equal number of input and output tokens. Unlike the supervised tokens $\boldsymbol{y}$, the tokens $\boldsymbol{v}$ and $\boldsymbol{q}$ serve as unsupervised conditional context, representing the model's comprehension of the input $\boldsymbol{x}$.


\noindent \textbf{Problem Definition.}
We explore knowledge distillation for MLLMs, aiming to distill small-scale MLLMs from powerful teacher MLLMs. Formally, given a teacher model distribution $P_T(\boldsymbol{y}|\boldsymbol{x})$ and a parameterized student distribution $P_S^\theta(\boldsymbol{y}|\boldsymbol{x})$, the knowledge distillation objective minimizes their distributional divergence on a dataset $\mathcal{D}=\{(\boldsymbol{x}^{(i)}, \boldsymbol{y}^{(i)})\}_{i=1}^{|\mathcal{D}|}$.

\noindent \textbf{Vanilla KD.}
Vanilla KD recognizes the autoregressive nature of language models and performs next-token alignment by combining ground-truth supervision with distribution matching. Given a query $\boldsymbol{x}$ and its corresponding ground-truth response sequence $\boldsymbol{y}_{1:L}^{\mathcal{D}}$, the objective is formalized as minimizing the forward KL divergence $\mathcal{L}_{\mathrm{kd}}(\theta)$ between $P_T$ and $P_S^{\theta}$ at each decoding step:
\begin{equation}
\mathcal{L}_{\mathrm{kd}}(\theta)=\mathop{\mathbb{E}}_{(\boldsymbol{x},\boldsymbol{y}_{1:L}^{\mathcal{D}}) \sim \mathcal{D}} \Biggl[ 
\sum_{k=1}^{L} 
D_{\text{KL}} (y_k \mid \boldsymbol{x}, \boldsymbol{y}_{<k}^{\mathcal{D}})
\Biggr],
\label{eq:sup_loss}
\end{equation}
where $D_{\text{KL}}(\cdot)$ represents the KL divergence between $P_T(\cdot)$ and $P_S^{\theta}(\cdot)$, and $\boldsymbol{y}_{<k}^{\mathcal{D}}$ denotes the ground-truth prefix tokens. This step-wise distillation transfers the teacher's generation preferences at each generation step.

\section{Framework of Align-TI}

Fig.~\ref{fig:pipeline} presents the overview of our proposed knowledge distillation framework, Align-TI, which is modeled in view of token interactions. It consists of two core components: (1) Instruction-aware Vision Alignment (IVA), which aligns visual tokens by incorporating instruction-aware importance weights to focus on salient visual regions. (2) Transition Probability Alignment (TPA), which aligns not only the initial token distributions conditioned on ground-truth data but also the transition probabilities to better imitate the teacher's autoregressive generation process.

\subsection{Instruction-aware Vision Alignment}

Unlike the conventional method~\citep{cai2024llavakd} that enforces uniform alignment across all visual tokens, our proposed IVA prioritizes alignment of the salient visual regions to mitigate the interference of redundant visual information. Specifically, we elaborate on the implementation of IVA from two aspects: identifying the most instruction-relevant visual focus and leveraging it for alignment.

\noindent \textbf{How to Identify Instruction-aware Visual Focus?}
Instruction-aware visual focus are inherently captured by the cross-modal attention mechanisms within MLLMs. Specifically, the instruction-to-vision attention map reflects how the model grounds textual queries onto visual features. However, the distribution of this visual focus evolves dynamically across layers, posing a challenge in selecting the optimal source. Visual token pruning methods~\citep{ye2025atp} similarly leverage these maps, but they often rely on empirical choices. To establish a principled selection criterion, we propose the \textbf{Instruction-Relevance Score (IRS)}. \textit{Our core intuition is that a layer performing effective semantic grounding should exhibit attention patterns that vary distinctively in response to different input instructions.}

\begin{definition}[Instruction-Relevance Score]
Let $\boldsymbol{\alpha}^{(l)}(\boldsymbol{x})$ denote the vectorized instruction-to-vision attention weights from the $l$-th layer for a given input $\boldsymbol{x}$. The Instruction-Relevance Score (IRS) for layer $l$ is defined as:
\begin{equation}
\mathrm{IRS}(l) = 1 - \mathbb{E}_{\substack{\boldsymbol{x}_1, \boldsymbol{x}_2 \\ \text{i.i.d.} \sim \mathcal{D}_x}} \left[ \cos \left( \boldsymbol{\alpha}^{(l)}(\boldsymbol{x}_1), \boldsymbol{\alpha}^{(l)}(\boldsymbol{x}_2)  \right) \right],
\end{equation}
where $\cos(\cdot, \cdot)$ denotes the cosine similarity function, and $\mathcal{D}_x$ is the distribution of input queries.
\end{definition}

\begin{figure}[ht]
  \centering
  \includegraphics[width=\columnwidth]{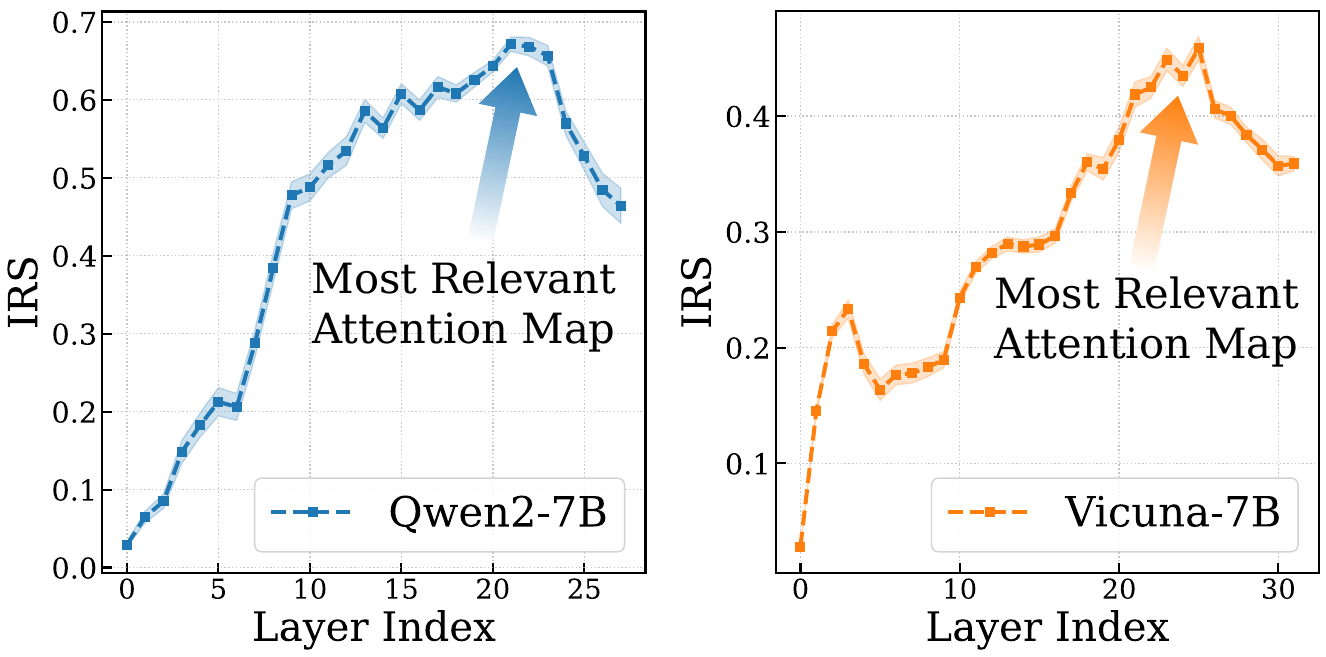}
  \vspace{-10pt}
  \caption{Analysis of IRS across layers with Qwen2-7B-based~\citep{bai2025qwen2} and Vicuna-7B-based~\citep{zheng2023judging} MLLMs.}
  \label{fig:layer}
\end{figure}

\noindent Fig.~\ref{fig:layer} visualizes the IRS across all layers for MLLMs based on Qwen2-7B and Vicuna-7B architectures. As illustrated in Figure~\ref{fig:layer}, both models exhibit a remarkably similar trend. It shows that shallow layers exhibit low IRS, indicating limited semantic grounding in early attention distributions. As depth increases, IRS increases, indicating that attention progressively sharpens and concentrates on instruction-relevant visual regions. However, this trend reverses in the few final layers, where the metric begin to decrease. We find that this is because the model starts integrating broader contextual information, causing the attention to lose its sharp focus on specific objects (see Appendix~\ref{appendix:iva} for visualization). Consequently, we select the layer $l^*$ with the maximal IRS to guide IVA, ensuring the supervision signal is derived from the most semantically focused representation.

\noindent \textbf{How to Align with Instruction-aware Visual Focus?}
Building upon the optimal layer $l^*$ identified by IRS, we formulate the IVA objective to transfer the teacher's visual extraction capability. Specifically, we extract the instruction-to-vision attention sub-matrix $\boldsymbol{A}_{i \rightarrow v} \in \mathbb{R}^{N_i \times N_v}$ from layer $l^*$, where $N_i$ and $N_v$ denote the numbers of instruction and visual tokens, respectively. We then aggregate the attention weights across the instruction dimension to derive the importance weight for each visual token $k$. These weights modulate the per-token alignment loss:
\begin{equation}
\label{eq:iva_loss}
\mathcal{L}_{\mathrm{iva}}(\theta) = \sum_{k=1}^{N_v} \frac{1}{N_i} \sum_{u=1}^{N_i} \boldsymbol{A}_{i\to v}(u,k) \cdot D_{\mathrm{KL}}(v_k \mid \boldsymbol{v}_{<k}),
\end{equation}
where $\frac{1}{N_i} \sum_{u=1}^{N_i} \boldsymbol{A}_{i\to v}(u,k)$ denotes the instruction-aware importance weight for the $k$-th
visual token. This formulation explicitly directs the student to allocate its representational capacity to the most salient visual information.

\noindent \textbf{Remark.}
The ultimate goal of KD is to train a student model to generate responses aligned with the teacher model. Although IVA aligns visual tokens rather than directly establishing relationships with response tokens, it enables the student model to imitate the teacher's processing of visual tokens. As a result, the hidden representations of the student model are implicitly optimized to be more effective for generating the target response.

\subsection{Transition Probability Alignment}
\label{sec:tpa}
In contrast to Vanilla KD, which aligns next-token predictions conditioned on ground-truth prefixes sampled from data $\mathcal{D}$, TPA aligns the token-to-token transition probability matrix. By emphasizing transition dynamics rather than only next-step predictions, TPA enables more effective transfer of sequential generation patterns and mitigates the growing discrepancy between teacher and student models during autoregressive decoding.

\noindent \textbf{Objective Derivation.}
Given a query $\boldsymbol{x}$ and a ground-truth prefix $\boldsymbol{y}_{<k}^{\mathcal{D}}$, our objective is to align not only the immediate next-token distribution but also the subsequent transition probabilities. We formalize this objective as the minimization of the following loss:
\begin{equation}
\begin{split}
\arg \min_{\theta}& \mathop{\mathbb{E}}_{(\boldsymbol{x}, \boldsymbol{y}_{1:L}^{\mathcal{D}}) \sim \mathcal{D}} \Biggl[
\sum_{k=1}^L \biggl(
\underbrace{D_{\mathrm{KL}}(y_k \mid \boldsymbol{x}, \boldsymbol{y}_{<k}^\mathcal{D})}_{\text{Initial State Align } \mathcal{L}_{\mathrm{kd}}(\theta)} \\
& \quad +
\underbrace{{\mathbb{E}}_{y_k \sim P_S^{\theta}} D_{\mathrm{KL}}(y_{k+1} \mid y_k, \boldsymbol{x}, \boldsymbol{y}_{<k}^\mathcal{D})}_{\text{Transition Probability Align } \mathcal{L}_{\mathrm{tpa}}(\theta)}
\biggr) \Biggr].
\end{split}
\end{equation}
The first term, $\mathcal{L}_{\mathrm{kd}}(\theta)$, is the objective of Vanilla KD, which aligns the initial state distribution at step $k$. The second term, $\mathcal{L}_{\mathrm{tpa}}(\theta)$, is our proposed objective, which aligns the one-step transition probability matrices:
\begin{equation}
P_T(y_{k+1} \mid y_k), P_S^{\theta}(y_{k+1} \mid y_k) \in \mathbb{R}^{|\mathcal{V}| \times |\mathcal{V}|},
\end{equation}
where $\mathcal{V}$ denotes the vocabulary set. These matrices capture token-to-token transition dependencies conditioned on $x$ and $\boldsymbol{y}_{<k}^{\mathcal{D}}$. Moreover, direct alignment of the full matrix is computationally infeasible due to the large vocabulary size $|\mathcal{V}|$. Fortunately, since initial probability distributions are highly long-tailed with most vocabulary entries having near-zero probability, aligning all matrix rows is unnecessary. We instead align transition probabilities conditioned on the student's on-policy distribution ($y_k \sim P_S^{\theta}$). Alternatively, conditioning on the off-policy teacher distribution ($y_k \sim P_T$) is possible, but it is less effective than conditioning on the student distribution. Specifically, the student distribution allows on-policy exploration of the student’s predictive space, facilitating correction of potential errors while maintaining computational efficiency by avoiding additional teacher forward passes.

\noindent \textbf{Remark.}
Consider the sequence decoding space defined over vocabulary $\mathcal{V}$. Vanilla KD aligns teacher-student distributions along $O(|\mathcal{V}|)$ generation paths by matching next-token probabilities. TPA expands this coverage to $O(|\mathcal{V}|^2)$ paths by additionally aligning the transition probability matrix, providing enhanced alignment scope. Detailed analysis is provided in Appendix~\ref{ana-scope:tpa}. This expanded scope enables supervision across more generation modes, achieving more exhaustive knowledge transfer. Furthermore, this expanded coverage increases the likelihood that student-generated trajectories during inference fall within the aligned distribution space, thereby mitigating the exposure bias caused by the train-test gap (described in Fig.~\ref{fig:intro} (Right)).

\noindent \textbf{Objective Estimation.}
We employ a Monte Carlo approach to estimate the expectation of $\mathcal{L}_{\mathrm{tpa}}(\theta)$. First, a forward pass through the student model yields the initial state probability distribution $P_S^{\theta}(y_k)$. We then sample a small set of $d$ candidate tokens $\{y_k^{(u)}\}_{u=1}^d$ from this distribution. Alignment is performed only on transition matrix rows associated with $\{y_k^{(u)}\}_{u=1}^d$, formalized as:
\begin{equation}\label{eq:tpa_estimation}
\mathcal{L}_{\mathrm{tpa}}(\theta) \simeq
\mathop{\mathbb{E}}_{(\boldsymbol{x},\boldsymbol{y}_{1:L}^{\mathcal{D}}) \sim \mathcal{D}} \sum_{k=1}^L \frac{1}{d}\sum_{u=1}^d D_{\mathrm{KL}}(y_{k+1} \mid y_k^{(u)}, \boldsymbol{x}, \boldsymbol{y}_{<k}^\mathcal{D}).
\end{equation}

\noindent \textbf{Parallelized Calculation.}
A naive implementation of Eq.~\ref{eq:tpa_estimation} would require $d$ separate forward passes on both the student model and the teacher model, incurring excessive computational cost. We propose an efficient parallel calculation method that computes these values in a single pass. Our approach begins by reorganizing the input response sequence as follows:
$\hat{\boldsymbol{y}} = \{y_{k-1}^{\mathcal{D}}, y_{k}^{(1)}, y_{k}^{(2)}, \dots, y_{k}^{(d)}\}_{k=1}^L$. During the forward pass of $\hat{\boldsymbol{y}}$, we apply a specially designed ribbon attention mask, as illustrated in Fig.~\ref{fig:ribbon}. Specifically, the mask ensures $y_{k}^{(u)}$ can only attend to the common context of $y_{k-1}$, but not to any other candidate $y_{k}^{(v)}$ ($v \neq u$). This strategy effectively creates $d$ parallel causal pathways, enabling simultaneous estimation of the required transition probabilities and significantly improving computational efficiency. Detailed efficiency analysis is provided in Appendix~\ref{sec:eff-tpa}.

\subsection{Overall Objective of Align-TI}

The overall objective of Align-TI integrates the standard SFT loss with our proposed distillation losses. The final objective $\mathcal{L}(\theta)$ is a summation of all components:
\begin{equation}
    \mathcal{L}(\theta)= \mathcal{L}_{\mathrm{sft}}(\theta)+\mathcal{L}_{\mathrm{iva}}(\theta)+\mathcal{L}_{\mathrm{kd}}(\theta)+\mathcal{L}_{\mathrm{tpa}}(\theta).
\end{equation}

\section{Experimental Results}
\label{sec:exp}
\noindent \textbf{Implementation Details.}
We utilize the Qwen2~\citep{bai2025qwen2} and Qwen3~\citep{yang2025qwen3} series as the LLMs for our student and teacher models. The teacher model comprises approximately 7-8B parameters, while the student model contains 1-2B parameters. The performance of teacher models is presented in Tab.~\ref{tab:teacher_perf}. Models within the same series are paired for distillation (e.g., Qwen2-7B serves as the teacher for Qwen2-0.5B/1.5B students). We follow the MobileVLM V2~\citep{mobilevlmv2} to organize our training data, with 1.2M captioning samples for pretraining and 2.4M mixed captioning and VQA samples for fine-tuning. Due to the limited learnable parameters, the KD is only adopted in the fine-tuning stage. The number of sampled tokens $d$ is set to 4. More implementation details and hyperparameters are illustrated in Appendix~\ref{sec:impl_details_training}. We mainly compare on six benchmarks to evaluate the multimodal understanding and VQA capabilities, more details about our evaluation benchmark are provided in Appendix~\ref{sec:impl_details_eval}.

\begin{table}[H]
    \vspace{-5pt}
    \caption{Performance of our Teacher Models.}
    \vspace{-5pt}
    \label{tab:teacher_perf}
    \centering
    \small
    \resizebox{\columnwidth}{!}{
    \begin{tabular}{lcccccccc}
        \toprule
        LLM & GQA & SQA & TextVQA & POPE & MME & MMB &AVG \\
        \midrule
        Qwen2-7B & 64.6 & 80.7 & 64.1 & 86.1 & 78.6 & 76.0 & 75.1 \\
        Qwen3-8B & 64.0 & 83.4 & 64.1 & 86.4 & 80.4 & 77.3 & 76.0 \\
        \bottomrule
    \end{tabular}
    }
    \vspace{-5pt}
\end{table}

\begin{table*}[ht]
\footnotesize
\setlength{\tabcolsep}{8pt}
\vspace{-2pt}
\caption{Comparison with state-of-the-art MLLMs. Our Align-TI establishes new state-of-the-art results among compact $\sim$1B and $\sim$2B parameter models, and demonstrates competitive performance with some larger models. $^\dagger$ denotes models obtained via MLLM distillation.}
\vspace{-5pt}
\label{tab:model_performance}
\centering
\resizebox{1.95\columnwidth}{!}{
\begin{tabular}{l|l|c|cccccc|c}
\toprule
Method         & LLM               & \#Params                  & GQA & SQA & TextVQA & POPE & MME & MMB & AVG \\ \midrule
LLaVA-1.5      & Vicuna-7B         & \multirow{5}{*}{$\sim$7B} & 62.0 & 66.8 & 58.2 & 85.9 & 75.5 & 64.3  & 68.8    \\
LLaVA-Next     & Vicuna-7B         &                           & 64.2 & 70.1 & 64.9 & 86.5 & 76.0 & 67.4  & 71.5    \\ 
MobileVLM V2   & Vicuna-7B         &                           & 62.6 & 74.8 & 62.3 & 85.3 & 78.0 & 69.2  & 72.0    \\
LLaVA-OV      & Qwen2-7B           &                           & 62.2 & -    & 84.5 & -    & 78.0 & 80.8  & -       \\ 
Qwen2.5-VL      & Qwen2.5-7B       &                           & 60.9 & 88.8 & 77.8 & 86.6 & 85.1 & 83.4  & 80.4    \\ \midrule
MiniCPM-V-2    & MiniCPM-2.4B      & \multirow{4}{*}{$\sim$3B}     & 52.1 & 76.3 & 73.2 & 86.3 & 70.5 & 68.5  & 71.2    \\
LLaVADI$^\dagger$        & MLLaMA-2.7B       &                           & 61.4 & 64.1 & 50.7 & 86.7 & 68.8 & 62.5  & 65.7    \\ 
MobileVLM V2   & MLLaMA-2.7B       &                           & 61.1 & 66.7 & 57.5 & 84.7 & 72.0 & 63.2  & 67.5    \\ 
TinyLLaVA      & Phi2-2.7B         &                           & 62.0 & 69.1 & 59.7 & 86.4 & 73.2 & 66.9  & 69.6    \\ \midrule
MoVE-KD$^\dagger$        & MobileLLaMA-1.4B       & \multirow{8}{*}{$\sim$2B} & 57.7 & 57.3 & 44.3 & 86.1 & 59.4 & 48.8  & 58.9    \\
Align-KD$^\dagger$       & MobileLLaMA-1.4B       &                           & 60.1 & 67.7 & 53.1 & \underline{87.0} & 65.2 & 57.5  & 65.1    \\
Mini-Gemini    & Gemma-2B          &                            & 60.7 & 63.1 & 56.2 & 85.6 & 67.0 & 59.8 & 65.4    \\
MoE-LLaVA      & Qwen1.5-1.8B      &                            & 61.5 & 63.1 & 48.0 & 87.0 & 64.6 & 59.7 & 64.0    \\
LLaVA-KD$^\dagger$       & Qwen1.5-1.8B       &                           & 62.3 & 64.7 & 53.4 & 86.3 & 69.1 & 64.0 & 66.6    \\
LLaVA-MoD$^\dagger$      & Qwen2-1.5B         &                           & 58.8 & 69.2 & 59.9 & \textbf{87.2}    & 69.2 & 68.9 & 68.9    \\
\rowcolor[RGB]{236,244,252}  Align-TI$^\dagger$     & Qwen2-1.5B      &   & \textbf{62.9}  & \underline{71.4} & \underline{65.1} & 86.1 & \textbf{75.6} & \underline{71.8} & \underline{72.2}    \\
\rowcolor[RGB]{236,244,252}  Align-TI$^\dagger$     & Qwen3-1.7B      &   & \underline{62.6}  & \textbf{76.5} & \textbf{67.1} & 86.6 & \underline{73.4} & \textbf{75.2} & \textbf{73.6}    \\ \midrule
SPHINX-Tiny    & TinyLLaMA-1.1B              & \multirow{5}{*}{$\sim$1B} & 58.0 & 21.5 & 57.8 & 82.2 & 63.1 & 56.6   & 56.5    \\
LLaVA-KD$^\dagger$       & Qwen1.5-0.5B            &                           & 59.6 & 60.6 & 49.9 & 85.9 & 64.5 & 60.1   & 63.4    \\
LLaVA-MoD$^\dagger$      & Qwen2-0.5B               &                           & 56.6 & \underline{61.1} & 57.1 & -    & 67.0 & 58.7  & -    \\
\rowcolor[RGB]{236,244,252}  Align-TI$^\dagger$    & Qwen2-0.5B             &                           & \underline{60.4} & 60.7 & \underline{59.9} & \underline{86.8} & \underline{68.9} & \underline{63.2}  & 66.7    \\
\rowcolor[RGB]{236,244,252}  Align-TI$^\dagger$    & Qwen3-0.6B             &                           & \textbf{61.2} & \textbf{68.4} & \textbf{64.1} & \textbf{86.9} & \textbf{70.0} & \textbf{67.6}  & \textbf{69.7}  \\ \bottomrule 
\end{tabular}
}
\end{table*}

\subsection{Main Results}
\noindent \textbf{Comparison with State-of-the-art MLLMs.}
As presented in Tab.~\ref{tab:model_performance}, we benchmark our models against state-of-the-art MLLMs, including both models trained from scratch and those derived via distillation. Our distilled Align-TI models achieve the best performance within the $\sim$1B and $\sim$2B parameter scales. Notably, Align-TI-2B surpasses substantially larger counterparts, outperforming LLaVA-1.5-7B and MobileVLM-V2-7B~\citep{mobilevlmv2} by relative $7.0\%$ and $2.2\%$, respectively. Furthermore, Align-TI-2B achieves a significant $4.8\%$ performance gain over LLaVA-MoD-2B~\citep{shu2024llavamod}, a superior MoE-based MLLM distillation baseline. These results demonstrate the efficacy of our distillation approach for transferring knowledge from large-scale MLLMs and developing high-performing small-scale MLLMs.

\begin{table}[t]
    \caption{Comparison with knowledge distillation strategies designed for LLM.}
    \vspace{-5pt}
    \label{tab:llm_cmp}
    \centering
    \resizebox{\columnwidth}{!}{
    \begin{tabular}{lccccccc}
        \toprule
        Model & GQA & SQA & TextVQA & POPE & MME & MMB & AVG \\
        \midrule
        FKL     & 59.3 & 59.7 & 59.2 & 86.2 & 65.0 & 60.4 & 65.0 \\
        MiniLLM & 59.4 & 59.9 & 58.7 & 85.5 & 65.6 & 57.9 & 64.5 \\
        JSD     & 57.4 & 58.9 & 57.9 & 85.9 & 66.2 & 56.1 & 63.7 \\
        GKD     & 55.4 & 58.5 & 57.3 & 85.6 & 61.5 & 57.9 & 62.7 \\
        \rowcolor[RGB]{236,244,252} Align-TI & \textbf{60.4} & \textbf{60.7} & \textbf{59.9} & \textbf{86.8} & \textbf{68.9} & \textbf{63.2} & \textbf{66.7} \\
        \bottomrule
    \end{tabular}
    }
\end{table}

\noindent \textbf{Comparison with Distillation Strategy Designed for LLMs.}
We further compare our method with distillation strategies specifically designed for LLMs, with results summarized in Tab.~\ref {tab:llm_cmp}. Classical distillation approaches for LLMs primarily employ diverse divergences, such as Forward KL (FKL), Jensen-Shannon Divergence (JSD) and Reverse KL in MiniLLM~\citep{gu2023minillm}. Our experiments show that FKL yields superior performance over other KL variants, which aligns with prior findings that the optimal divergence is task-dependent~\citep{agarwal2024policy, xu2024llavadi}. Additionally, GKD~\citep{agarwal2024policy} exhibits an average performance degradation of 2.3 relative to FKL. This may be attributed to GKD aligning on student-generated on-policy responses, which can lead to incorrect answers, particularly in more challenging multi-modal scenarios. Nevertheless, all these LLM-centric strategies are substantially outperformed by Align-TI, as the gap between LLM and MLLM is significant.

\noindent \textbf{Efficiency Analysis.}
We analyze the training overhead of our approach in Tab.~\ref{tab:train_eff}. It demonstrates that IVA incurs negligible additional computational overhead, with the majority of the cost arising from TPA. \textit{Moreover, IVA can be seamlessly combined with TPA at almost zero cost.} Overall, our complete Align-TI method increases the training time by $1.4\times$ compared to Vanilla KD, which is substantially more efficient than GKD's $2.7\times$ overhead. This demonstrates that Align-TI achieves its performance gains at a modest training cost.

\begin{table}[ht]
    \vspace{-5pt}
    \caption{Training Efficiency.}
    \vspace{-7pt}
    \label{tab:train_eff}
    \centering
    \resizebox{\columnwidth}{!}{
    \begin{tabular}{lccccc}
        \toprule
        \multirow{2}{*}{Metrics} & \multirow{2}{*}{Vanilla KD} & \multirow{2}{*}{GKD} & \multicolumn{3}{c}{Align-TI} \\
        \cmidrule(lr){4-6}
        & & & IVA & TPA & TPA+IVA \\
        \midrule
        Training Time (H) & 355 & 962 & 359 & 504 & 509 \\
        Memory (GiB)      & 70.6 & 76.8 & 70.7 & 75.3 & 75.6 \\
        \bottomrule
    \end{tabular}
    }
    \vspace{-5pt}  
\end{table}

\subsection{Ablation Study}
\label{sec:exp-layer}
\noindent \textbf{Component Analysis.} To evaluate the contributions of IVA and TPA, Tab.~\ref{tab:component} presents an ablation study on their individual and combined effects. When neither IVA nor TPA is employed, the baseline model achieves an average performance of 64.3. In contrast, the combined integration of TPA and IVA yields an average performance of 66.7, which represents an improvement of 2.4 over the baseline. When applied separately, TPA and IVA improve the baseline by 2.1 and 0.8, respectively, confirming the efficacy of each module. Notably, the performance gain from TPA is larger than that from IVA. This observation aligns with their distinct mechanisms: TPA imposes an explicit constraint by directly aligning output distributions, whereas IVA operates indirectly by matching latent feature representations for response generation. Moreover, as illustrated in Tab.~\ref{tab:train_eff}, the training time overhead of TPA is significantly higher than that of IVA, while IVA can be integrated with TPA at nearly no additional cost.

\begin{table}[ht]
    \caption{Ablation Study on IVA and TPA.}
    \vspace{-7pt}
    \label{tab:component}
    \centering
    \resizebox{\columnwidth}{!}{
    \begin{tabular}{ccccccccc}
        \toprule
        IVA & TPA & GQA & SQA & TextVQA & POPE & MME & MMB & AVG \\
        \midrule
        {\color{gray}\ding{55}} & {\color{gray}\ding{55}} & 57.6 & 59.0 & 59.1 & 86.2 & 66.1 & 57.9 & 64.3 \\
        \checkmark & {\color{gray}\ding{55}} & 59.6 & 58.0 & \textbf{61.3} & 86.5 & 66.8 & 58.3  & 65.1 \\
        {\color{gray}\ding{55}} & \checkmark & 60.3 & \textbf{61.0} & 59.6 & 86.5 & 68.1 & 63.0 & 66.4 \\
        \checkmark & \checkmark & \textbf{60.4} & 60.7 & 59.9 & \textbf{86.8} & \textbf{68.9} & \textbf{63.2} & \textbf{66.7} \\
        \bottomrule
    \end{tabular}
    }
    \vspace{-5pt}
\end{table}

\begin{table}[ht]
    \small
    \centering
    \caption{Compare with visual token alignment with uniform weights~\citep{cai2024llavakd}.}
    \vspace{-7pt}    
    \label{tab:iva-uniform-cmp}
    \resizebox{1.0\linewidth}{!}{
    \setlength{\tabcolsep}{1.3mm}
    \begin{tabular}{l|cccccc|c}
    \toprule
    Method & GQA & SQA & TextVQA & POPE & MME & MMB & AVG \\
    \midrule
    Uniform  & 59.4 & 57.3 & 61.0 & 86.3 & 65.1 & 58.0 & 64.5 \\
    \rowcolor[RGB]{236,244,252}  Ours     & \textbf{59.6} & \textbf{58.0} & \textbf{61.3} & \textbf{86.5} & \textbf{66.8} & \textbf{58.3} & \textbf{65.1} \\
    \bottomrule
    \end{tabular}
    }
    \vspace{-5pt}
\end{table}

\textbf{Comparison with Uniform Alignment.}
To demonstrate the efficacy of our proposed IVA, we benchmark it against the uniform alignment strategy from~\citep{cai2024llavakd}, which assigns equal weights to all visual tokens. As shown in Tab.~\ref{tab:iva-uniform-cmp}, IVA outperforms uniform alignment on all six benchmarks, achieving an average improvement of 0.6. This highlights that IVA's instruction-aware weighting mechanism facilitates a more effective and targeted alignment.

\noindent \textbf{Design of Important Weights.}
Tab.~\ref{tab:layer-selection-iva} presents our investigation into the optimal layer depth for extracting importance weights. By evaluating five equidistant layers, we find that optimal performance is achieved with the 21st layer, located approximately three-quarters of the model's depth. Notably, this layer coincides with the model region exhibiting the highest IRS. In contrast, weights from layers with low instruction relevance degrade model performance. This finding underscores that IVA's effectiveness originates from its ability to focus student model's limited capability on instruction-salient regions while filtering out the impact of redundant visual tokens. Furthermore, we compare the effects of using importance weights derived from the student and teacher models, as detailed in Tab.~\ref {tab:student-teacher-iva}. The results reveal that aligning with the teacher's attention focus yields an average performance improvement of 0.4. This improvement can be attributed to the fact that the student model, with its limited capacity and lack of SFT, does not yet possess a robust ability to accurately focus on instruction-relevant regions.

\begin{table}[ht]
    \caption{Comparison of importance weights extracted from different layer depths for IVA.}
    \vspace{-5pt}
    \label{tab:layer-selection-iva}
    \centering
    \small
    \resizebox{\columnwidth}{!}{
    \setlength{\tabcolsep}{1.3mm}
    \begin{tabular}{lccccccc}
        \toprule
        Layer & GQA & SQA & TextVQA & POPE & MME & MMB & AVG \\
        \midrule
        \phantom{0}0 (0/4) & 58.1 & 56.9 & 58.9 & 86.3 & 66.3 & 55.8 & 63.7 \\
        \phantom{0}7 (1/4) & 55.4 & 56.1 & 40.8 & 85.1 & 62.6 & 53.0 & 58.8 \\
        14 (2/4)           & 58.3 & 57.1 & 59.6 & 85.8 & 66.0 & 56.7 & 63.9 \\
        \rowcolor[RGB]{236,244,252}
        21 (3/4)           & \textbf{59.6} & 58.0 & \textbf{61.3} & 86.5 & 66.8 & \textbf{58.3} & \textbf{65.1} \\
        27 (4/4)           & 58.8 & \textbf{58.2} & 59.3 & \textbf{86.8} & \textbf{67.1} & 57.5 & 64.6 \\
        \bottomrule
    \end{tabular}
    }
    \vspace{-5pt}
\end{table}

\begin{table}[ht]
    \caption{Comparison of importance weights extracted from teacher and student models for IVA.}
    \vspace{-7pt}
    \label{tab:student-teacher-iva}
    \centering
    \small
    \resizebox{\columnwidth}{!}{
    \setlength{\tabcolsep}{1.3mm}
    \begin{tabular}{lccccccc}
        \toprule
        Method & GQA & SQA & TextVQA & POPE & MME & MMB & AVG \\
        \midrule
        Student & 59.3 & 57.5 & 61.1 & 86.3 & 65.4 & \textbf{58.3} & 64.7 \\
        \rowcolor[RGB]{236,244,252}
        Teacher & \textbf{59.6} & \textbf{58.0} & \textbf{61.3} & \textbf{86.5} & \textbf{66.8} & \textbf{58.3} & \textbf{65.1} \\
        \bottomrule
    \end{tabular}
    }
    \vspace{-5pt}
\end{table}

\begin{figure}[t!]
\centering
\begin{minipage}{0.48\columnwidth}
    \centering
    \includegraphics[width=\linewidth]{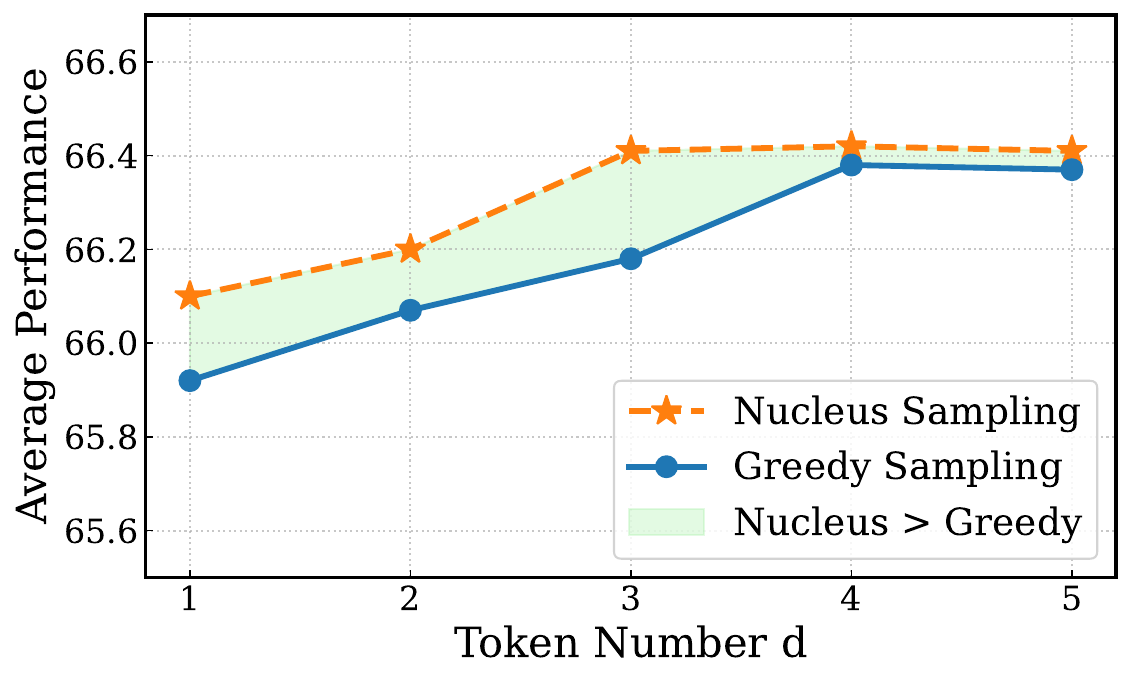}
    \vspace{-20pt}
    \caption{Ablation study on TPA design choices: comparing different sampling strategies and sampled token number $d$.}
    \label{fig:sampling}
\end{minipage}
\hfill
\begin{minipage}{0.48\columnwidth}
    \centering
    \includegraphics[width=\linewidth]{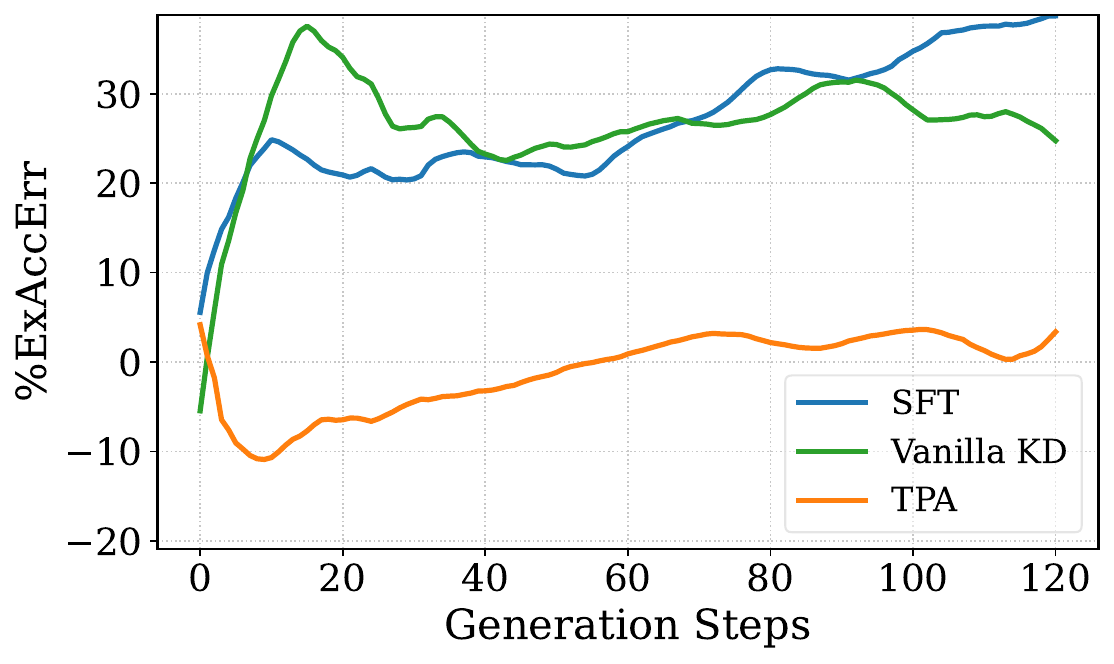}
    \vspace{-20pt}
    \caption{Evolution of $\mathrm{\%ExAccErr}$ across generation steps, illustrating TPA's effect on mitigating exposure bias.}
    \label{fig:exaccerr}
\end{minipage}
\vspace{-10pt}
\end{figure}

\noindent \textbf{Impact of Different TPA Designs.}
We compare two sampling strategies for TPA: (i) greedy sampling, which selects the top-$d$ most probable tokens, and (ii) nucleus sampling, which stochastically draws $d$ tokens from the student's predictive distribution. Fig.~\ref{fig:sampling} illustrates the model performance as a function of sampled token number $d$. For both strategies, performance improves as $d$ increases, eventually plateauing around $d=4$. This trend aligns with the expectation that sampling the high-probability part of the output distribution is adequate since language model output distributions are typically long-tailed. More importantly, nucleus sampling consistently outperforms greedy sampling in the low-$d$ regime. We attribute this to the diversity inherent in nucleus sampling, which encourages the student to learn a broader range of state transitions. In contrast, greedy sampling focuses on the model's most confident, high-probability tokens, which often overlap with the ground truth, thereby providing redundant supervision.


\subsection{Analysis on IVA and TPA}
\begin{figure}[ht]
  \centering
  \includegraphics[width=0.98\linewidth]{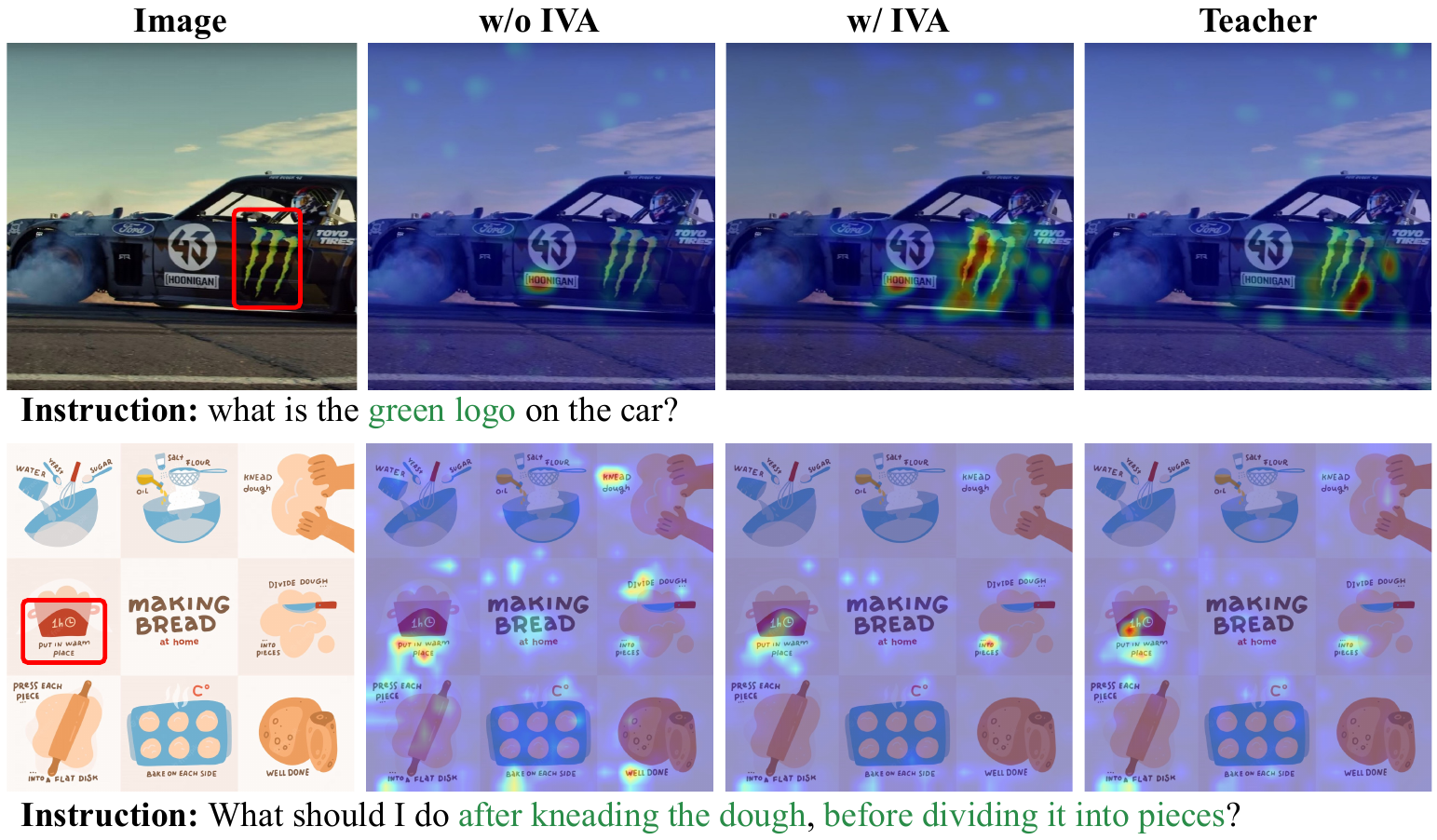}
  \vspace{-5pt}    
  \caption{Qualitative analysis of IVA. IVA enhances student attention by correcting misdirected focus (top row) and sharpening diffuse attention maps into precise ones (bottom row).}
  \label{fig:iva_quality}
  \vspace{-12pt}
\end{figure}

\textbf{Analysis of IVA on Enhancing Visual Focus.}
We qualitatively examine how IVA strengthens a student model’s ability to attend to instruction-relevant visual regions. As illustrated in Fig.~\ref{fig:iva_quality}, we visualize the instruction-to-vision attention maps for the student model (with and without IVA) alongside the teacher model. These maps are sourced from the layer exhibiting the highest IRS. We observe that equipping the student with IVA makes its attention patterns more closely align with the teacher's. Specifically, we identify two primary improvements: \textbf{(1) Focus correction}: Without IVA, the student may incorrectly attend to unrelated objects. For instance, when asked about a ``green logo," it focuses on an entirely different logo (top row). IVA helps redirect its attention to the correct target. \textbf{(2) Focus sharpening}: Even when the student localizes the correct general area without IVA, its attention can be dispersed across irrelevant regions. IVA refines this into a concentrated map that closely follows the teacher’s precise focus (bottom row). These findings demonstrate that IVA effectively distills the teacher's ability to extract visual information.

\noindent \textbf{Analysis of TPA on Mitigating Exposure Bias.}
Fig.\ref{fig:exaccerr} depicts the analysis of $\mathrm{\%ExAccErr}$, a metric for assessing exposure bias with definition and calculation details outlined in Appendix~\ref{appendix:exacccerr}. When exposure bias is eliminated, the teacher and student models generate the same prediction distribution across different prefixes, resulting in a $\mathrm{\%ExAccErr}$ being zero. As shown in Fig.~\ref{fig:exaccerr}, models trained with SFT and Vanilla KD both exhibit significant exposure bias, with their $\mathrm{\%ExAccErr}$ values quickly rising and then stabilizing at around $30\%$. In contrast, the model distilled through our TPA approach exhibits $\mathrm{\%ExAccErr}$ within the range of $(-10\%, 10\%)$. Moreover, the $\mathrm{\%ExAccErr}$ even becomes negative in the early stages, indicating that the gap between student and teacher is smaller in the condition of a prefix generated by the student model. This finding strongly suggests that TPA successfully forces the student to learn the teacher's underlying transition dynamics, thereby aligning their output distributions across different contexts and effectively reducing exposure bias.

\begin{figure}[ht]
    \centering
    \vspace{-3pt}
    \includegraphics[width=0.6\columnwidth]{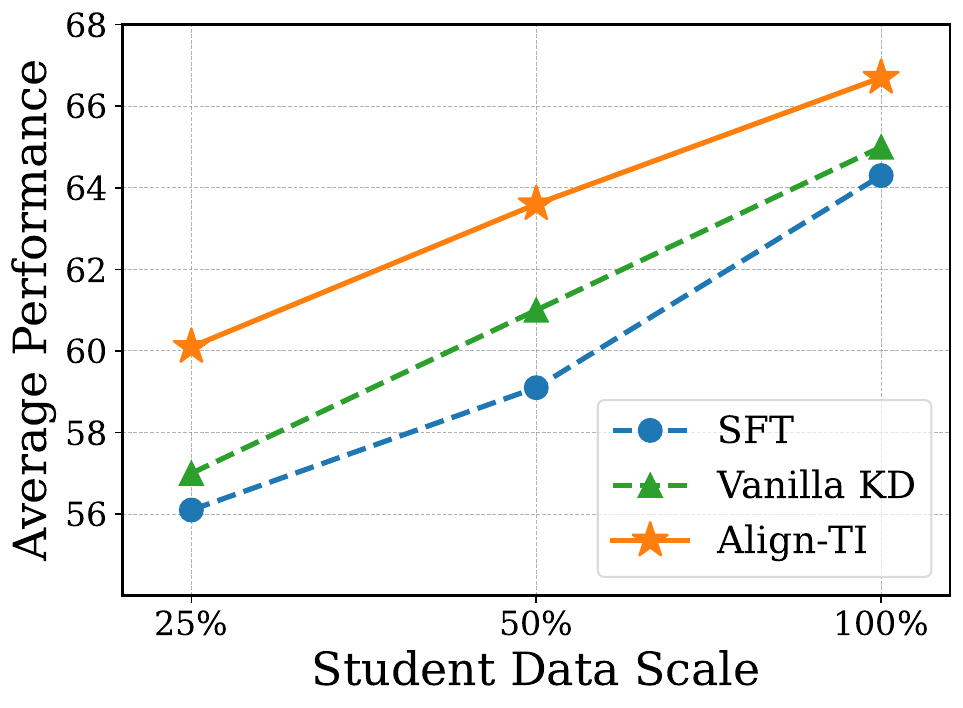}
    \vspace{-5pt}
    \caption{The scaling law for student training data.}
    \label{fig:scaling_data}
    \vspace{-10pt}
\end{figure}

\subsection{Scaling Analysis}
\noindent \textbf{Data Scaling.} Fig.~\ref{fig:scaling_data} presents the performance of SFT, Vanilla KD and Align-TI as a function of training data size, with Qwen2-0.5B as the LLM of student. All three approaches demonstrate consistent improvements as the amount of training data increases. Notably, Align-TI consistently outperforms Vanilla KD across all data scales. Furthermore, the results indicate that SFT and Vanilla KD perform similarly when trained on either a limited dataset ($25\%$) or the full dataset ($100\%$), whereas Align-TI delivers substantial gains in both settings. This highlights two key strengths of Align-TI: (1) it facilitates highly effective knowledge transfer even in data-scarce scenarios, and (2) it continues to distill supplementary knowledge when data is abundant, enabling further performance gains.

\noindent \textbf{Teacher Scaling.}
Tab.~\ref{tab:teacher-scale} presents the impact of teacher model size on Align-TI, using Qwen3-1.7B/4B/8B and Qwen3-0.6B as the LLM of teacher and student. Increasing the teacher size from 2B to 4B results in a notable average performance improvement, as indicated by 1.4 on SQA and 1.3 on GQA. However, further increasing the scale to 8B results in only marginal gains, with the average performance improving by a mere 0.1. This suggests that the benefits of a larger teacher exhibit diminishing returns, likely constrained by the representation capacity of the student model, a phenomenon also observed in~\citep{mirzadeh2020improved}.
\vspace{-2pt}

\begin{table}[ht]
    \small
    \centering
    \vspace{-5pt}
    \caption{Scaling analysis of teacher model size.}
    \vspace{-7pt}
    \label{tab:teacher-scale}
    \resizebox{1.0\linewidth}{!}{
    \setlength{\tabcolsep}{1.3mm}
    \begin{tabular}{c|cccccc|c}
    \toprule
    Teacher Size & GQA & SQA & TextVQA & POPE & MME & MMB & AVG \\
    \midrule
    $\sim$2B	     & 60.2	& 67.4 & 63.5 & 86.4 & \textbf{70.5}	& 67.5 & 69.3 \\
    $\sim$4B	     & \textbf{61.5}	& \textbf{68.8} & 63.7 & 86.2 & 69.7	& 67.5 & 69.6 \\
    \rowcolor[RGB]{236,244,252} $\sim$8B	     & 61.2	& 68.4 & \textbf{64.1} & \textbf{86.9} & 70.0	& \textbf{67.6} & \textbf{69.7} \\
    \bottomrule
    \end{tabular}
    }
\end{table}

\noindent \textbf{Model Architecture.}
To validate the effectiveness of Align-TI across diverse architectures, we conduct additional experiments using MobileLLaMA-1.4B~\citep{chu2024mobilevlm} as the student model, and MLLM with Vicuna-7B~\citep{chiang2023vicuna} as the teacher. As shown in Tab.~\ref{tab:teacher-arch}, Vanilla KD yields merely a 0.2 average improvement over SFT, while exhibiting a 1.5 performance drop on MME. In contrast, our method achieves an average gain of 2.1 relative to SFT, and achieves the best performance on all benchmarks. These results demonstrate the robustness of Align-TI across various model architectures.

\begin{table}[ht]
\small
\centering
\caption{Distillation performance comparison on MobileLLaMA-1.4B as the student LLM.}
\vspace{-7pt}
\resizebox{1.0\linewidth}{!}{
\setlength{\tabcolsep}{1.3mm}
\begin{tabular}{l|cccccc|c}
\toprule
Method & GQA & SQA & TextVQA & POPE & MME & MMB & AVG \\
\midrule
SFT	             & 58.4	& 64.8 & 60.6 & 85.6 & 66.6	& 56.2 & 65.4 \\
Vanilla KD	     & 58.9	& 65.6 & 61.4 & 86.0 & 65.1	& 56.6 & 65.6 \\
\rowcolor[RGB]{236,244,252} Align-TI    & \textbf{60.4} & \textbf{67.7} & \textbf{61.9} & \textbf{86.9} & \textbf{67.5}	& \textbf{60.7} & \textbf{67.5} \\
\bottomrule
\end{tabular}
\label{tab:teacher-arch}
}
\end{table}

\section{Conclusion}
This paper introduces Align-TI, a novel token-level knowledge distillation framework for transferring knowledge from large-scale to parameter-efficient MLLMs. Viewing distillation through the lens of token interactions, we analyze vision-instruction token interactions and intra-response token interactions, and propose two components: Instruction-aware Vision Alignment (IVA) and Transition Probability Alignment (TPA). IVA aligns visual tokens on instruction-aware salient regions to learn the teacher's visual information extraction capability, while TPA distills token-to-token transition probabilities to transfer the dynamics of autoregressive generation. Experiments demonstrate Align-TI’s effectiveness in distilling MLLMs.

\section*{Broader Impact}
This paper presents Align-TI, which focuses on the efficiency and accessibility of Multimodal Large Language Models (MLLMs). By proposing Align-TI, a framework that effectively distills large-scale models into compact ones, our work contributes to reducing the computational resources and energy consumption required for MLLM inference. This facilitates the deployment of advanced multimodal capabilities on resource-constrained edge devices, thereby democratizing access to this technology.

\bibliography{main}
\bibliographystyle{icml2026}

\clearpage
\newpage
\appendix
\onecolumn
\section*{Appendix}
In Sec.~\ref{sec:related_work}, we review existing works relevant to this study. Sec.~\ref{sec:impl_details} provides more implementation details of this study, including training details, benchmark details, comparison methods, the calculation of some metrics and the implementation details of TPA. Additional technical details for our proposed IVA and TPA are presented in Sec.~\ref{appendix:iva} and Sec.~\ref{appendix:tpa}, respectively. Sec.~\ref{sec:add-exp} presents additional experiments. Sec.~\ref{sec:limitation} discusses the limitations and potential future directions of this work. Finally, we provide several case studies in Sec.~\ref{sec:add_analysis}.

\section{Related Work}
\label{sec:related_work}
\noindent \textbf{Multimodal Large Language Models.}
The success of large language models, driven by self-supervised next-token prediction, has significantly advanced multimodal learning by unifying vision and language modalities within the LLM framework. To achieve this, aligning visual and textual representations is essential. Key alignment strategies include: Flamingo \citep{alayrac2022flamingo}, which integrates visual features via cross-attention adapters; BLIP-2~\citep{li2023blip}, employing a Querying Transformer for visual-language pre-alignment; and LLaVA~\citep{liu2023visual}, which demonstrated that a simple MLP layer suffices for effective modality alignment and is widely adopted in subsequent works~\citep{bai2025qwen2,chen2024far}. Beyond LLMs, MLLMs also rely on pre-trained vision encoders to process vision input. Thus, \citep{li2024mini} introduces a more powerful vision encoder, and \citep{huang2025hires} supports higher-resolution inputs. Furthermore, MLLM capabilities are expanding beyond text generation to tasks like segmentation~\citep{liu2025seg} and detection~\citep{shen2025vlmr1}. Recent works~\citep{zhao2025r1omini,r-4b} introduce the reasoning capabilities into the MLLMs. In addition, considering the substantial computational demands of MLLMs, significant efforts~\citep{chu2024mobilevlm, zhou2024tinyllava, zhang2025llava} are also focused on improving model efficiency. This work also focuses on compact MLLMs and develops a more powerful knowledge distillation method for transferring knowledge in large-scale MLLMs.

\noindent \textbf{Knowledge Distillation for LLM.}
Recent years have witnessed remarkable successes in LLMs trained on extensive datasets with numerous parameters. However, their substantial computational requirements limit deployment in resource-constrained scenarios, motivating extensive research into model compression via KD~\citep{hinton2015distilling}. KD transfers knowledge from powerful teacher models to compact student models, categorized into black-box (output-only access) and white-box (access to intermediate features and logits) paradigms~\citep{xu2024survey}. White-box KD, with access to intermediate features and logits, shows superior knowledge transfer capabilities compared to black-box KD methods that synthesize data through the teacher model~\citep{kim2016sequence,guha2025openthoughts,openr1}. This study primarily focuses on white-box distillation.

Early efforts in LLM distillation centered on refining the distillation objective. For instance, MiniLLM~\citep{gu2023minillm}, $f$-Distill~\citep{wen2023fdistill} and DistiLLM~\citep{ko2024distillm} proposed using reverse KL divergence, $f$-divergence and skew KL divergence, respectively, to better align the student's output distribution with the teacher's. More recent advancements have focused on transferring more complex forms of knowledge. GKD~\citep{agarwal2024policy} enables the student to learn from the teacher's rationale on self-generated mistakes. PromptKD~\citep{kim2024promptkd} pioneered the use of prompt tuning to adapt the teacher, making its knowledge more student-friendly. Furthermore, RLKD~\citep{xu2025distilling} introduced a reinforcement learning framework guided by a novel reward model, allowing the student LLM to internalize the teacher's complex, multi-branch reasoning pathways.

\noindent \textbf{Knowledge Distillation for MLLM.}
Since LLMs serve as the backbone of MLLMs, KD techniques developed for LLMs provide a foundational basis. However, their direct application is suboptimal, as MLLMs introduce the additional knowledge from a vision encoder and preserving cross-modal alignment. The exploration of MLLM distillation is still in its early stages. A recent study by LLaVADI~\citep{xu2024llavadi} revealed that features from intermediate layers, attention mechanisms, and token relationships are ineffective for MLLM distillation. Moreover, LLaVA-KD~\citep{cai2024llavakd} established a framework incorporating both multimodal content and relational distillation. To address limitations in student capacity, LLaVA-MoD~\citep{shu2024llavamod} enhances the student's representational power by integrating a Mixture-of-Experts (MoE) architecture. Align-KD~\citep{feng2025alignkd} focuses on modeling the cross-modal alignment process during distillation. Additionally, some research concentrates on distilling vision encoders, such as MoVE-KD~(Cao et al., 2025), which utilizes multi-teacher distillation to efficiently compress the vision encoder. Furthermore, our work proposes a novel MLLM distillation framework centered on the interactions in the prefilling and decoding stages.

\begin{table*}[ht]
\centering
\caption{Configuration for training teacher model.}
\label{tab:hyper}
\begin{tabular}{@{}
              p{0.25\textwidth}
              >{\centering\arraybackslash}p{0.23\textwidth}
              >{\centering\arraybackslash}p{0.23\textwidth}@{}}
\toprule
Configuration & Pretraining & Fine-tuning \\ \midrule
Trainable components & VL-projector & VL-projector, LLM \\
Epochs               & 1            & 1 \\
Batch size           & 256          & 128 \\
Learning rate        & $1 \times 10^{-3}$ & $2 \times 10^{-5}$ \\
LR scheduler type    & Cosine       & Cosine \\
Optimizer            & AdamW        & AdamW \\
Weight decay         & 0            & 0 \\
Warmup ratio         & 0.03         & 0.03 \\
BF16                 & True         & True \\
Model max length     & 2048         & 2048 \\
Fine-tuning type     & Full         & Full \\
Model parallelism    & ZeRO-2       & ZeRO-2 \\ 
\bottomrule
\end{tabular}
\end{table*}

\section{Implementation Details}
\label{sec:impl_details}
\subsection{Training Details}
\label{sec:impl_details_training}
\subsubsection{Training Details for Teacher Models}
\label{sec:impl_details_training_teacher}
In this study, we employ token-level knowledge distillation to train small-scale MLLMs. To maintain architectural consistency with the target small-scale models and facilitate effective knowledge transfer, we follow established protocols~\citep{liu2023visual} for training large-scale teacher MLLMs. Our teacher models utilize Qwen2-7B and Qwen3-8B as the backbone language models, and following recent best practices, we employ the SigLIP-B/14~\citep{zhai2023sigmoid} as the visual encoder. The vision-language projector consists of a two-layer MLP with GeLU activation.

We adopt the two-stage training paradigm from LLaVA~\citep{liu2023visual}: (1) Pretraining stage: Models are trained on the LLaVA1.5-558k caption dataset~\citep{liu2023visual} for one epoch with a learning rate of $10^{-3}$ and batch size of 256. (2) Fine-tuning stage: Models are trained on the LLaVA-mix-665k dataset~\citep{liu2023visual}, which combines caption and VQA data, for one epoch with a learning rate of $2 \times 10^{-5}$ and batch size of 128. Both stages employ the AdamW optimizer~\citep{loshchilov2017decoupled} with cosine decay learning rate scheduling and warmup, utilizing full parameter fine-tuning. Our implementation builds upon the open-source LLaVA codebase and is conducted on 8 NVIDIA A100 GPUs. Detailed training hyperparameters are provided in Tab.~\ref{tab:hyper}.

\begin{table*}[ht]
\centering
\caption{Detailed description for data adopted to train the student models.}
\label{tab:student_data}
\begin{tabular}{llcl}
\toprule
Stage                        & Datasets      & \#Samples & Description  \\ \midrule
Pretraining                  & ShareGPT4V-PT~\citep{chen2024sharegpt4v} & 1.2M      & Caption      \\ \midrule
\multirow{9}{*}{Fine-tuning} & VSR~\citep{liu2023vsr}           & 13K       & VQA          \\
                             & SQA~\citep{lu2022sqa}           & 13K       & VQA          \\
                             & Text-VQA~\citep{singh2019textvqa}      & 35K       & VQA          \\
                             & VIGC~\citep{wang2024vigc}          & 37K       & VQA          \\
                             & IConQA~\citep{lu2021iconqa}        & 107K      & VQA          \\
                             & Visual Dialog~\citep{das2017visual} & 123K      & Conversation \\
                             & COCO~\citep{chen2015microsoft}          & 592K      & Caption      \\
                             & ShareGPT4V~\citep{chen2024sharegpt4v}    & 665K      & Mixed        \\
                             & SBU~\citep{ordonez2011im2text}           & 844K      & Caption      \\ \midrule
Total                        & -             & 3.6M      & -            \\ \bottomrule
\end{tabular}
\end{table*}

\subsubsection{Training Details for Student Models}
\label{sec:impl_details_training_student}
Aligned with the teacher models' two-stage training strategy, student models employ identical hyperparameter configurations for learning rate, training epochs, and optimizer. Two key distinctions exist: First, student models utilize more compact large language architectures, specifically Qwen2-0.5B/1.5B and Qwen3-0.6B/1.7B. Second, during fine-tuning, we introduce auxiliary supervision from the teacher model via knowledge distillation. The training objective thus combines standard supervised fine-tuning loss with our proposed distillation loss.

To address the constrained capacity of compact models, which typically require expanded training corpora, we adopt established small-model training methodologies~\citep{mobilevlmv2}. This motivates our use of an augmented dataset containing 3.6M samples: 1.2M captioning samples during pretraining and 2.4M mixed captioning and VQA samples during fine-tuning. Data sources and functional allocations are detailed in Tab.~\ref{tab:student_data}. All experiments were conducted across 16 NVIDIA H20 GPUs.

\subsection{Benchmark Details}
\label{sec:impl_details_eval}
Our comprehensive evaluation encompasses six carefully curated benchmarks designed to assess diverse visual-language understanding and generation capabilities. The key characteristics of each benchmark are detailed below:

\noindent\textbf{GQA} \citep{hudson2019gqa} (\textit{Question Answering on Image Scene Graphs}): A VQA benchmark designed for real-world visual reasoning and compositional question answering, containing 12,578 samples for evaluation.

\noindent\textbf{SQA} \citep{lu2022sqa} (\textit{Scientific question answering}): 
A VQA benchmark focusing on scientific tables and text with diverse reasoning types, containing 4,241 samples for evaluation.

\noindent\textbf{TextVQA} \citep{singh2019textvqa} (\textit{Text Visual Question Answering}): A VQA benchmark for visual reasoning based on text in images, containing 5,000 samples for evaluation.

\noindent\textbf{POPE} \citep{li2023pope} (\textit{Polling-based Object Probing Evaluation}): A Hallucination detection benchmark focusing systematically evaluating object hallucination tendencies, containing 8,910 samples for evaluation.

\noindent\textbf{MME} \citep{fu2024mme} (\textit{Multimodal Model Evaluation}): Comprehensive evaluation suite covering measures both perception
and cognition abilities on a total of 14 subtasks with 2,374 manually curated samples.

\noindent\textbf{MMB}~\citep{liu2024mmbench} (\textit{MultiModal Benchmark}): Large-scale multi-choice VQA benchmark containing questions requiring advanced reasoning across 20 task categories, with 4,377 English questions for MMB.

\subsection{Details of Comparison Methods}
\label{sec:impl_details_methods}
We evaluate our Align-TI method by distilling MLLMs at two different scales: $\sim$1B and $\sim$2B parameters. Our comparison encompasses both models of similar parameter counts and larger-scale models to demonstrate the efficiency and effectiveness of our approach.

For models with comparable parameter counts ($\sim$1B and $\sim$2B), we compare against a comprehensive set of baselines spanning different training paradigms. These include models trained end-to-end from scratch, such as SPHINX-Tiny~\citep{liu2024sphinx}, Mini-Gemini~\citep{li2024mini}, and MoE-LLaVA~\citep{lin2024moe}. Additionally, we compare with models developed using specialized knowledge distillation techniques for MLLMs, including MoVE-KD~\citep{cao2025movekd}, Align-KD~\citep{feng2025alignkd}, LLaVA-KD~\citep{cai2024llavakd} and LLaVA-MoD~\citep{shu2024llavamod}.

To further demonstrate the parameter efficiency of our method, we extend our comparison to larger-scale models. This includes $\sim$3B parameter models such as TinyLLaVA~\citep{zhou2024tinyllava}, MobileVLM V2~\citep{mobilevlmv2}, LLaVADI~\citep{xu2024llavadi} and MiniCPM-V-2~\citep{minicpmv}, as well as $\sim$7B parameter models including LLaVA-1.5~\citep{llava15}, LLaVA-Next~\citep{liu2024llavanext}, LLaVA-OV~\citep{li2024llavaov} and Qwen2.5-VL~\citep{bai2025qwen2}.

\subsection{Details of Analysis on Exposure Bias}
\label{sec:impl_details_exp_bias}
\subsubsection{Training-time and Test-time Accumulated Error}
\label{sec:accerr-intro}
In Fig.~\ref{fig:intro}, we visualize the training-time and test-time accumulated errors to reveal the expanding gap with increasing generation length. This section provides the detailed computation. To ensure analysis over sufficiently long sequences, we construct an evaluation set $\mathcal{D}^{e}$ by randomly sampling 1K samples from the original training set where response length exceeds 100 tokens. These samples are subsequently removed from the training data, ensuring $\mathcal{D}^{e}$ follows the same distribution as the training set. The student model analyzed here is trained using Vanilla KD.

The training-time accumulated error $E_{\text{train}}(l)$ measures the cumulative divergence under teacher forcing, where the model is conditioned on the ground-truth prefix $y_{<t}$ at each generation step:
\begin{align}
    E_{\text{train}}(l) &= \sum_{t=1}^l \mathbb{E}_{(\boldsymbol{x}, \boldsymbol{y}) \sim \mathcal{D}^{e}} \left[ D_{\mathrm{KL}}( P_T \| P_S^\theta)(y_t \mid \boldsymbol{x}, \boldsymbol{y}_{<t}) \right].
\end{align}
In contrast, the test-time accumulated error $E_{\text{test}}(l)$ simulates realistic autoregressive inference by conditioning on the model's own generated prefix:
\begin{align}
    E_{\text{test}}(l) &= \sum_{t=1}^l \mathbb{E}_{\boldsymbol{x} \sim \mathcal{D}_x^{e}, \, \boldsymbol{y}_{<t} \sim P_S^\theta} \left[ D_{\mathrm{KL}}( P_T \| P_S^\theta)(y_t \mid \boldsymbol{x}, \boldsymbol{y}_{<t}) \right].
\end{align}
The train-test gap illustrated in Fig.~\ref{fig:intro} corresponds to the difference $E_{\text{test}}(l) - E_{\text{train}}(l)$, which directly quantifies the performance degradation caused by error propagation during inference.

\subsubsection{Excess Accumulated Error}
\label{appendix:exacccerr}

\begin{definition}[Excess Accumulation Error]\label{def:exposure_bias}
Given a target distribution $P_T$ and a parameterized student model $P_S^\theta$, the Excess Accumulated Error ($\%\mathrm{ExAccErr}_{\le}(l)$)~\citep{arora2022exposure} quantifying exposure bias over sequences is formally defined as:
\begin{equation}\label{eq:accerr}
\%\mathrm{ExAccErr}_{\le}(l) = \frac{R(l) - E(l)}{E(l)} \times 100\%,
\end{equation}
where $R(l)$ denotes the accumulated regret of imitating the teacher's generation logic up to $l$ time steps, and $E(l)$ is the baseline error conditioned on the oracle context sampled from teacher distribution $P_T$. A value near zero implies mitigation of exposure bias.
\end{definition}

The excess accumulated error is estimated using the 1K dataset $\mathcal{D}_x^{e}$ described in Sec.~\ref{sec:accerr-intro}. Here, $R(l)$ represents the accumulated teacher-student error up to generation step $l$, conditioned on low-quality prefixes sampled from the student model $P_S^{\theta}$, and is estimated using KL divergence:
\begin{align}
    R(l)=\sum_{t=1}^l \mathbb{E}_{\boldsymbol{x}\sim \mathcal{D}^{e}_x, \boldsymbol{y}_{<t} \sim P_S^\theta} \left[ D_{\mathrm{KL}}( P_T \| P_S^\theta)(y_t \mid \boldsymbol{x}, \boldsymbol{y}_{<t}) \right]
\end{align}
Similarly, $E(l)$ denotes the baseline teacher-student error up to generation step $l$, conditioned on oracle contexts sampled from the teacher distribution $P_T$:
\begin{align}
    E(l)=\sum_{t=1}^l \mathbb{E}_{\boldsymbol{x}\sim \mathcal{D}^{e}_x, \boldsymbol{y}_{<t} \sim P_T} \left[ D_{\mathrm{KL}}( P_T \| P_S^\theta)(y_t \mid \boldsymbol{x}, \boldsymbol{y}_{<t}) \right]
\end{align}
Thus, $\%\mathrm{ExAccErr}_{\le}(l)$ quantifies the relative error induced by exposure bias. Under ideal conditions where exposure bias is effectively mitigated, the student model should exhibit nearly identical distribution gaps regardless of the source model generating the response, resulting in $\%\mathrm{ExAccErr}_{\le}(l) \rightarrow 0$. 
Notably, due to the uncertain relationship between $R(l)$ and $E(l)$, $\%\mathrm{ExAccErr}_{\le}(l)$ can assume negative values. A negative value indicates that the distribution gap between student and teacher becomes particularly small when conditioned on student-generated responses, while a larger gap exists when conditioned on teacher-generated oracle responses. This phenomenon suggests the persistent presence of exposure bias.

\subsection{Calculation Details of IRS.}
\label{sec:impl_details_irs}
The Instruction-Relevance Score (IRS) is formally defined as the expected cosine similarity between the vision token importance vectors extracted from attention maps for two different inputs. We estimate this expectation by constructing a set of 1K input pairs, each containing two different queries. Our empirical results demonstrate that the IRS is a stable metric that converges rapidly with a modest number of sample pairs.

\subsection{Implementation Details of TPA.}
\begin{figure}[h!]
  \centering
  \includegraphics[width=0.4\linewidth]{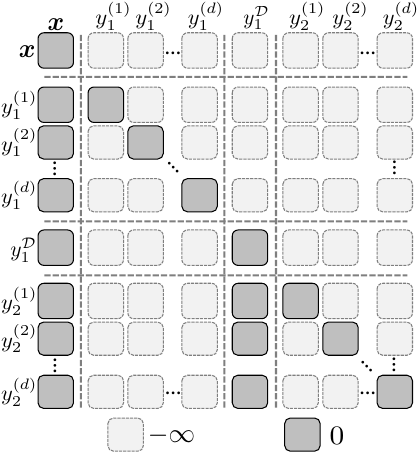}
  \caption{Visualization of the ribbon attention mask used for parallelizing the TPA computation.}
  \label{fig:ribbon}
\end{figure}

Algorithm~\ref{alg:tpa} outlines the procedure for Transition Probability Alignment (TPA), providing further implementation details. Moreover, to enable efficient parallel computation (discussed in Sec.~\ref{sec:tpa}), we utilize a ribbon attention mask, as visualized in Figure~\ref{fig:ribbon}.

\begin{algorithm*}[h]
\caption{Transition Probability Alignment (TPA)}
\label{alg:tpa}
\begin{algorithmic}[1]
\STATE \textbf{Input:} Frozen teacher model $P_T(\boldsymbol{y}|\boldsymbol{x})$
\STATE Student model $P_S^{\theta}(\boldsymbol{y}|\boldsymbol{x})$ with learnable parameters $\theta$.
\STATE Training dataset $\mathcal{D}$
\STATE Hyperparameters: number of sampled tokens $d$, learning rate $\eta$
\REPEAT
    \STATE Sample a mini-batch $\mathcal{B} \sim \mathcal{D}$.
    \STATE Initialize total losses $\mathcal{L}_{\mathrm{kd}}(\theta) \gets 0$ and $\mathcal{L}_{\mathrm{tpa}}(\theta) \gets 0$.
    \FOR{each $(\boldsymbol{x}, \boldsymbol{y}^{\mathcal{D}})$ in $\mathcal{B}$}
        \STATE Initialize augmented sequence $\hat{\boldsymbol{y}} \gets \emptyset$.
        \STATE Perform single forward pass of $P_S^{\theta}$ on $(\boldsymbol{x}, \boldsymbol{y}^{\mathcal{D}})$.
        \FOR{$k = 1$ to $|\boldsymbol{y}^{\mathcal{D}}|$}
            \STATE Sample $d$ candidate tokens $\{y_{k+1}^{(i)}\}_{i=1}^{d} \sim P_S(\cdot \mid \boldsymbol{x}, \boldsymbol{y}^{\mathcal{D}}_{<k})$ 
            \STATE Concatenate tokens: $\hat{\boldsymbol{y}} \gets \hat{\boldsymbol{y}} \circ y_k^{\mathcal{D}} \circ \big[y_{k+1}^{(1)}, \dots, y_{k+1}^{(d)}\big]$
        \ENDFOR
        \STATE Construct the ribbon attention mask $M$ for $\hat{\boldsymbol{y}}$  \COMMENT{Ensures parallel causal paths, Fig.~\ref{fig:ribbon}}
        \STATE Perform single forward passes of $P_T$ and $P_S^{\theta}$ on $(\boldsymbol{x}, \hat{\boldsymbol{y}})$ using attention mask $M$.
        \STATE Compute the initial state alignment loss (Vanilla KD): \COMMENT{Eq.~\ref{eq:sup_loss}}
        \STATE $\mathcal{L}_{\mathrm{kd}}(\theta) \gets \mathcal{L}_{\mathrm{kd}}(\theta) +  \sum_{k=1}^{|\boldsymbol{y}^{\mathcal{D}}|}D_{\mathrm{KL}}\big(P_T \parallel P_S\big)(y_k \mid \boldsymbol{x}, \boldsymbol{y}^{\mathcal{D}}_{<k})$
        \STATE Compute the transition probability alignment loss: \COMMENT{Eq.~\ref{eq:tpa_estimation}}
        \STATE $\mathcal{L}_{\mathrm{tpa}}(\theta) \gets \mathcal{L}_{\mathrm{tpa}}(\theta) +  \sum_{k=1}^{|\boldsymbol{y}^{\mathcal{D}}|}\frac{1}{d} \sum_{i=1}^{d} D_{\mathrm{KL}}\big(P_T \parallel P_S\big)(y_{k+1} \mid y_k^{(i)},\boldsymbol{x}, \boldsymbol{y}^{\mathcal{D}}_{<k})$
    \ENDFOR
\UNTIL{convergence}
\STATE \textbf{Return:} Distilled student model $P_S^{\theta}(\boldsymbol{y}|\boldsymbol{x})$
\end{algorithmic}
\end{algorithm*}

\section{Additional Details for IVA}
\label{appendix:iva}

\subsection{Visualization Analysis for Instruction-to-Vision Attention Map.}
\begin{figure*}[h]
    \centering
    \includegraphics[width=\linewidth]{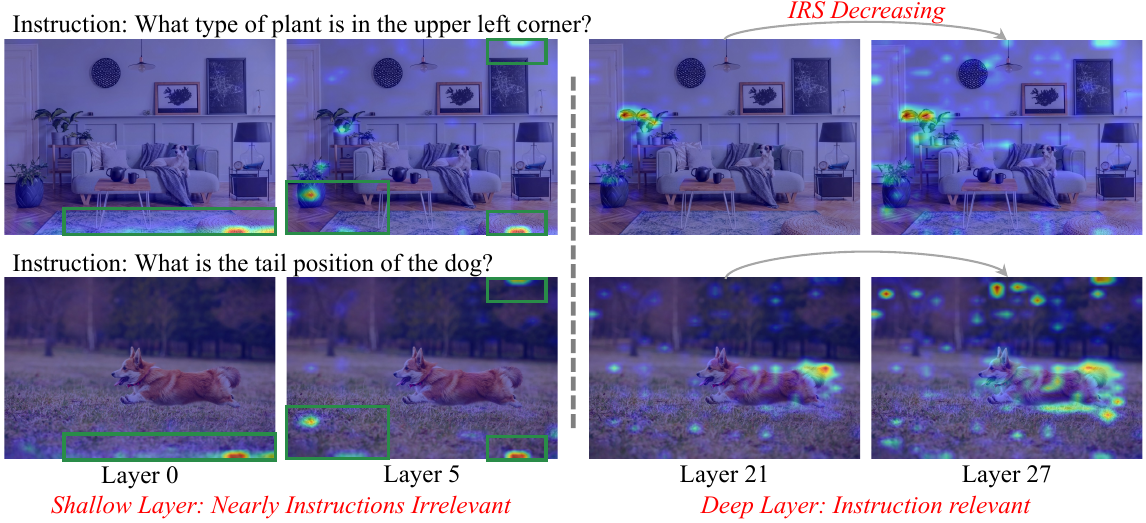}
    \caption{Qualitative analysis of instruction-to-vision attention map evolution across layers. In shallow layers, the attention is largely instruction-agnostic, with different instructions causing the model to attend to similar visual regions (highlighted in gree boxes). In deep layers, the attention maps become highly instruction-specific, with model focusing on instruction-relevant visual regions.}
    \label{fig:qualitative_attention}
\end{figure*}

As quantitatively illustrated in Fig.~\ref{fig:layer}, the IRS varies significantly across different layers of the teacher model. Specifically, IRS is relatively low in the early layers, decreases to its maximum in the middle layers, and then gradually decreases in the deeper layers. To qualitatively interpret this behavior, we visualize visual token importance maps across layers in Fig.~\ref{fig:qualitative_attention}. Our analysis reveals a clear evolution of the model's attention mechanism. Initially, in the shallow layers (e.g., Layers 0 and 5), the model demonstrates instruction-agnostic behavior, with attention maps focusing on general salient patterns regardless of the instruction. This corresponds to an initial phase of low-level feature extraction, resulting in low IRS. As information propagates to the middle layers (e.g., Layer 21), the attention transitions to an instruction-specific mode, sharply focusing on semantically relevant regions for each task. This semantic filtering process causes a significant drop in IRS to their maximum. Interestingly, towards the final layers (e.g., from Layer 21 to 27), the focused attention begins to diffuse, re-incorporating contextual information from surrounding areas. This final stage of contextual refinement, essential for generating rich responses, leads to a subsequent decreasing in IRS.

\section{Additional Details for TPA}
\label{appendix:tpa}

\begin{figure*}[htbp]
    \centering
    \includegraphics[width=0.8\linewidth]{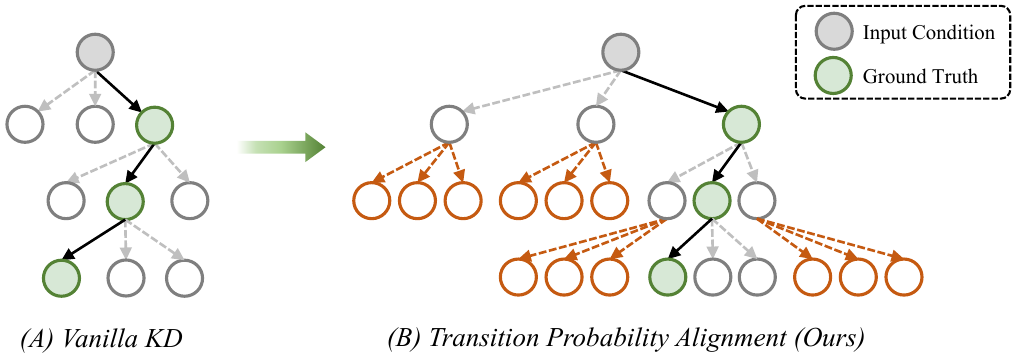}
    \caption{Visualization of alignment spaces achieved by Vanilla KD and Transition Probability Alignment. Starting from the root node, probability distributions at internal nodes and leaf nodes are aligned to transfer knowledge from teacher to student models. Each non-leaf node has children spanning the entire vocabulary $\mathcal{V}$.}
    \label{fig:space_expand}
\end{figure*}

\subsection{Derivation of Alignment Scope Expanded in TPA}
\label{ana-scope:tpa}
We analyze the alignment scope by conceptualizing the generation process as a tree structure where each node represents a token state and edges represent transitions between states.

Vanilla KD aligns next-token probabilities conditioned on ground-truth prefixes. At each timestep $k$, it minimizes $D_{\mathrm{KL}}(P_T \| P_S^{\theta})(y_k | x, \boldsymbol{y}_{<k}^{\mathcal{D}})$ over all $|\mathcal{V}|$ possible next tokens. However, the alignment follows a single trajectory determined by the ground-truth sequence $\boldsymbol{y}_{<k}^{\mathcal{D}}$. In the tree representation, Vanilla KD aligns all child nodes at each level but only traverses one path from root to leaf. For a sequence of length $L$, this results in alignment over $L \times |\mathcal{V}|$ nodes, yielding $O(|\mathcal{V}|)$ path coverage as illustrated in Fig.~\ref{fig:space_expand} (A).

TPA extends this alignment by incorporating transition probability matching through $\mathbb{E}_{y_k \sim P_S^{\theta}} D_{\mathrm{KL}}(P_T \| P_S^{\theta})(y_{k+1} | y_k)$. Rather than expanding only from ground-truth tokens, TPA samples candidate tokens and aligns transitions from each sampled state. This corresponds to aligning a $|\mathcal{V}| \times |\mathcal{V}|$ transition matrix at each timestep, where both predecessor and successor tokens span the entire vocabulary. For a sequence of length $L$, TPA aligns $|\mathcal{V}| + (L-1) \times |\mathcal{V}|^2$ nodes, achieving $O(|\mathcal{V}|^2)$ path coverage as shown in Fig.~\ref{fig:space_expand} (B).

\subsection{Discussion on Training Efficiency of TPA}
\label{sec:eff-tpa}
Tab.~\ref{tab:train_eff_tpa} presents the training efficiency analysis of our proposed TPA, specifically evaluating the computational overhead in terms of forward propagation frequency. We consider a dataset consisting of $N$ training samples with an average response sequence length of $L$, and denote the number of sampled tokens as $d$. As shown in the comparison, Vanilla KD requires $N$ forward pass for both the student and teacher models per epoch. Utilizing a parallelized calculation strategy with carefully designed attention masks, our proposed TPA necessitates $2N$ forward passes for the student and $N$ for the teacher. This mechanism ensures that each token attends solely to its valid prefix, enabling efficient batch processing.

Without this parallelization, the computational cost becomes prohibitively expensive, scaling to $dLN + N$ for the student and $dLN$ for the teacher, as the output distribution must be computed iteratively for each sampled token. Consequently, our parallelized approach significantly reduces the overhead, ensuring that Align-TI maintains a training efficiency comparable to standard KD methods while achieving superior alignment performance.

\begin{table}[htbp]
\centering
\caption{Comparison of computational overhead in terms of forward passes per epoch. Here, $N$ denotes the number of training samples, $L$ represents the average response sequence length, and $d$ indicates the number of sampled tokens.}
\label{tab:train_eff_tpa}
\begin{tabular}{lcc}
\toprule
\multirow{2}{*}{Method} & \multicolumn{2}{c}{Number of Forward Passes} \\
\cmidrule(lr){2-3}
 & Student Model & Teacher Model \\
\midrule
Vanilla KD & $N$ & $N$ \\
TPA (w/ Parallelization) & $2N$ & $N$ \\
TPA (w/o Parallelization) & $dLN + N$ & $dLN$ \\
\bottomrule
\end{tabular}
\end{table}

\subsection{Discussion on impact of TPA on Sequence-level Alignment.}
In this section, we aim to gain deeper insight by examining how our proposed TPA relates to the ideal objective of sequence-level alignment. The ultimate goal of knowledge distillation is to minimize the KL divergence between the teacher's and the student's full sequence distributions:
\begin{align}\label{eq:seqkd}
\mathop{\min}_{\theta} \mathbb{E}_{\boldsymbol{x} \sim \mathcal{D}_x} \left[ D_{\text{KL}}\left(P_T \parallel P_S^{\theta}\right)(\boldsymbol{y}_{1:L} \mid \boldsymbol{x}) \right]
\end{align}
where $L$ is the sequence length and $\mathcal{D}_x$ denotes the set of input problems. Our analysis reveals that, compared to Vanilla KD, TPA promotes this sequence-level alignment. We present evidence from both theoretical and experimental perspectives.

\noindent \textbf{Theoretically.} 
Strictly optimizing the formulation in Eq.~\ref{eq:seqkd} necessitates alignment within a joint probability space of complexity $O(|V|^L)$. In practice, this space is computationally intractable and highly sparse, with numerous combinations being semantically meaningless or contextually irrelevant. As discussed in Sec.~\ref{ana-scope:tpa}, Vanilla KD simplifies this objective by performing alignment in an $O(|V|)$ space. In contrast, TPA operates within an $O(|V|^2)$ space. This implies that TPA exposes the student to richer structural patterns and transition dynamics during training. Such alignment provides a more superior approximation of the full sequence distribution compared to Vanilla KD, thereby facilitating the optimization of Eq.~\ref{eq:seqkd}.

\noindent \textbf{Empirically.} 
A primary motivation for TPA is to mitigate the Exposure Bias arising from the training-test distribution shift. We employ the Excess Accumulated Error metric ($\mathrm{\%ExAccErr}$) to quantify the severity of this bias. Fig.~\ref{fig:exaccerr} illustrates the trajectory of $\mathrm{\%ExAccErr}$ as the number of generation steps increases. The results indicate that the model distilled via TPA maintains a remarkably low error rate (ranging between 0 and 10\%), which is significantly lower than the approximately 30\% observed with Vanilla KD. This substantial reduction suggests that the cumulative error between the student and teacher is effectively suppressed during autoregressive generation. Consequently, the sequences generated by the student exhibit higher fidelity to the teacher's distribution, empirically confirming that Eq.~\ref{eq:seqkd} is better optimized under TPA.

Moreover, better alignment with Eq.~\ref{eq:seqkd} implies that the student model internalized more teacher's underlying continuous generative logic, thereby improving its ability to capture long-range dependencies and transcending the limitations of simple token-level mimicry.

\section{Additional Experiments}
\label{sec:add-exp}

\subsection{Loss Contribution Analysis.}
Table \ref{tab:loss} illustrates the impact of each loss term on final performance, providing a clearer understanding of their respective contributions. The overall objective of our proposed Align-TI consists of four components: Supervised Fine-Tuning (SFT) loss $\mathcal{L}_{\mathrm{sft}}$, Instruction-aware Vision Alignment (IVA) loss $\mathcal{L}_\mathrm{iva}$, Vanilla KD loss $\mathcal{L}_\mathrm{kd}$ and Transition Probability Alignment (TPA) loss $\mathcal{L}_\mathrm{tpa}$. Starting from the baseline with only $\mathcal{L}_{\mathrm{sft}}$, adding $\mathcal{L}_{\mathrm{iva}}$ or $\mathcal{L}_{\mathrm{kd}}$ individually yields average improvements of 0.8 and 0.7, respectively, while combining the two results in a larger gain of 1.3. When initiating from a Vanilla KD configuration, incorporating $\mathcal{L}_{\mathrm{iva}}$ and $\mathcal{L}_{\mathrm{tpa}}$ enhances performance by 0.6 and 1.4, respectively. Applying both losses together achieves a total enhancement of 1.7. Notably, the absence of $\mathcal{L}_{\mathrm{kd}}$ in this setup only slightly decreases performance by 0.1, suggesting that the other components can effectively compensate for its omission.

\begin{table*}[h]
\centering
\small
\caption{Impact of each loss term to the final performance.}
\label{tab:loss}
\begin{tabular}{l|cccccc|c}
\toprule
Loss                                                                          & GQA & SQA & TextVQA & POPE & MME & MMB & AVG \\ \midrule
$\mathcal{L}_{\mathrm{sft}}$                                                  & 57.6 & 59.0 & 59.1 & 86.2 & 66.1 & 57.9 & 64.3     \\
$\mathcal{L}_{\mathrm{sft}}+\mathcal{L}_\mathrm{iva}$                         & 59.6 & 58.0 & \textbf{61.3} & 86.5 & 66.8 & 58.3  & 65.1     \\
$\mathcal{L}_{\mathrm{sft}}+\mathcal{L}_\mathrm{kd}$                          & 59.3 & 59.7 & 59.2 & 86.2 & 65.0 & 60.4 & 65.0     \\
$\mathcal{L}_{\mathrm{sft}}+\mathcal{L}_\mathrm{iva}+\mathcal{L}_\mathrm{kd}$ & 59.8  & 59.4 & 60.5 & 86.5 & 66.3	& 61.3	& 65.6   \\
$\mathcal{L}_{\mathrm{sft}}+\mathcal{L}_\mathrm{kd}+\mathcal{L}_\mathrm{tpa}$ & 60.3	& \textbf{61.0}	& 59.6	& 86.5	& 68.1	& 63.0	& 66.4     \\
$\mathcal{L}_{\mathrm{sft}}+\mathcal{L}_\mathrm{tpa}+\mathcal{L}_\mathrm{iva}$& 60.3 	   & 60.4	& 60.6	& 86.4	& 68.5	& 63.1	& 66.6     \\
$\mathcal{L}_{\mathrm{sft}}+\mathcal{L}_\mathrm{iva}+\mathcal{L}_\mathrm{kd}+\mathcal{L}_\mathrm{tpa}$ & \textbf{60.4} & 60.7 & 59.9 & \textbf{86.8} & \textbf{68.9} & \textbf{63.2} & \textbf{66.7}     \\ \bottomrule
\end{tabular}
\end{table*}

\subsection{Details of Figure 1 (Right).}
\label{appendix:fig1-details}
Fig.~\ref{fig:intro-performance} (Right) provides a bar chart comparison to highlight the performance differences between our proposed Align-TI, SFT, and Vanilla KD. The data for this visualization is drawn from the comprehensive results presented in Tab.~\ref{tab:llm_cmp} and~\ref{tab:component}. For a more direct examination, we reproduce the exact numerical values in Tab.~\ref{tab:fig1_right}.

\begin{table}[H]
\centering
\small
\caption{Comparison with standard SFT and Vanilla KD.}
\label{tab:fig1_right}
\begin{tabular}{l|cccccc|c}
\toprule
Model & GQA & SQA & TextVQA & POPE & MME & MMB & AVG \\
\midrule
SFT & 57.6 & 59.0 & 59.1 & 86.2 & 66.1 & 57.9 & 64.3 \\
Vanilla KD     &59.3 & 59.7 & 59.2 & 86.2 & 65.0 & 60.4 & 65.0 \\
Align-TI & \textbf{60.4} & \textbf{60.7} & \textbf{59.9} & \textbf{86.8} & \textbf{68.9} & \textbf{63.2} & \textbf{66.7} \\
\bottomrule
\end{tabular}
\end{table}

\subsection{Performance Comparison Between Teacher and Student.}
In Tab.~\ref{tab:teacher_student_comparison}, we compare the performance of the teacher model with that of its distilled student counterparts. The results show that student models trained via distillation achieve performance comparable to that of the teacher. Moreover, increasing the student model's parameter count from 1B to 2B leads to consistent improvements, narrowing the gap with the teacher. Notably, both the 1B and 2B student models exhibit lower hallucination rates than the teacher. On TextVQA, the student models perform on par with or even surpass the teacher. We attribute this to the benchmark's relative simplicity and the teacher's potential over-parameterization in this context. A similar trend of improvement is also observed in LLaVA-MoD~\citep{shu2024llavamod}. However, on more complex benchmarks such as SQA, MME, and MMB, a substantial performance gap remains.

\begin{table*}[h]
\centering
\small
\caption{Comparison between teacher models and distilled student models}
\label{tab:teacher_student_comparison}
\begin{tabular}{ll|cccccc|c}
\toprule
Type & LLM & GQA & SQA & TextVQA & POPE & MME & MMB & AVG \\
\midrule
Teacher & Qwen2-7B & 64.6 & 80.7 & 64.1 & 86.1 & 78.6 & 76.0 & 75.1 \\ \midrule
\multirow{2}{*}{Student} & Qwen2-1.5B & 62.9 & 71.4 & 65.1 & 86.1 & 75.6 & 71.8 & 72.2 \\
                         & Qwen2-0.5B & 60.4 & 60.7 & 59.9 & 86.8 & 68.9 & 63.2 & 66.7 \\
\midrule \midrule
Teacher & Qwen3-8B & 64.0 & 83.4 & 64.1 & 86.4 & 80.4 & 77.3 & 76.0 \\  \midrule
\multirow{2}{*}{Student} & Qwen3-1.7B & 62.6 & 76.5 & 67.1 & 86.6 & 73.4 & 75.2 & 73.6 \\
                         & Qwen3-0.6B & 61.2 & 68.4 & 64.1 & 86.9 & 70.0 & 67.6 & 69.7 \\
\bottomrule
\end{tabular}
\end{table*}

\subsection{Inference Efficiency Analysis}
We evaluate the inference efficiency of our Align-TI-2B model against the much larger LLaVA-1.5-7B. As shown in Tab.~\ref{tab:infer_eff}, Align-TI-2B achieves $1.7\times$ faster first-token generation, $1.9\times$ higher decoding throughput, and consumes only $4.8$\,GiB of peak memory. These characteristics underscore its suitability for deployment on resource-constrained edge devices.

\begin{table}[H]
\centering
\small
\caption{Inference Efficiency Analysis.}
\label{tab:infer_eff}
\begin{tabular}{lcc}
\toprule
Metrics & LLaVA-1.5 & Align-TI \\
\midrule
Params (B) & 7 & 2 \\
Peak Memory (GiB) & 14.0 & 4.8 \\
Time to First Token (ms) & 90 & 57 \\
Throughput (token/s) & 33.8 & 64.8 \\
AVG Performance (\%) & 68.8 & 73.6 \\
\bottomrule
\end{tabular}
\end{table}

\subsection{Comparison with Vanilla KD under a Similar Computational Budget}
Our proposed method, Align-TI, requires more training time per epoch compared to the Vanilla KD baseline (509 hours vs. 355 hours). A valid concern is whether the performance gains of Align-TI stem from this increased training time rather than its algorithmic design. To ablate the impact of the computational budget, we conducted an additional experiment where we extended the training of Vanilla KD to match the total training time of our method. Specifically, we increased the training for Vanilla KD from 1.0 epoch to 1.5 epochs. This raised its total training time to 533 hours, which is comparable to the 509 hours required for a single epoch of Align-TI. The comparative results are presented in Table~\ref{tab:time_comparison}. The results show that extending Vanilla KD's training yields only marginal performance gains, with its average score increasing from 65.0 to 65.3. In contrast, Align-TI achieves a significantly higher average score of 66.7 within a similar timeframe. This experiment strongly suggests that the performance superiority of our method is not a mere consequence of a larger training budget but is primarily attributed to its more effective and efficient algorithmic design.

\begin{table*}[h]
\centering
\small
\caption{Performance comparison with Vanilla KD under a similar training time budget.}
\label{tab:time_comparison}
\begin{tabular}{lcccccccc}
\toprule
Method & Training Time (H) & GQA & SQA & TextVQA & POPE & MME & MMB & AVG \\
\midrule
Vanilla KD (1 epoch) & 355 & 59.3 & 59.7 & 59.2 & 86.2 & 65.0 & 60.4 & 65.0 \\
Vanilla KD (1.5 epochs) & 533 & 59.5 & 59.5 & 59.4 & 86.4 & 66.1 & 60.6 & 65.3 \\
Align-TI (1 epoch) & 509 & \textbf{60.4} & \textbf{60.7} & \textbf{59.9} & \textbf{86.8} & \textbf{68.9} & \textbf{63.2} & \textbf{66.7} \\
\bottomrule
\end{tabular}
\end{table*}

\section{Limitations and Future Work.}
\label{sec:limitation}
Due to limited computational resources, we validate the effectiveness of Align-TI only on image–text benchmarks. However, we believe Align-TI can be effectively extended to other modalities (e.g., video), as such tasks also produce token-level outputs, which can be modeled by Align-TI's objective. Exploring its potential in continuous spaces represents a promising direction, enabling application to diverse models such as unified frameworks and latent reasoning architectures. Moreover, exploring the vision-language alignment via distillation of vision-language projector could be a promising avenue.

\section{Case Study}
\label{sec:add_analysis}
\begin{figure*}[htbp]
    \centering
    \includegraphics[width=0.9\linewidth]{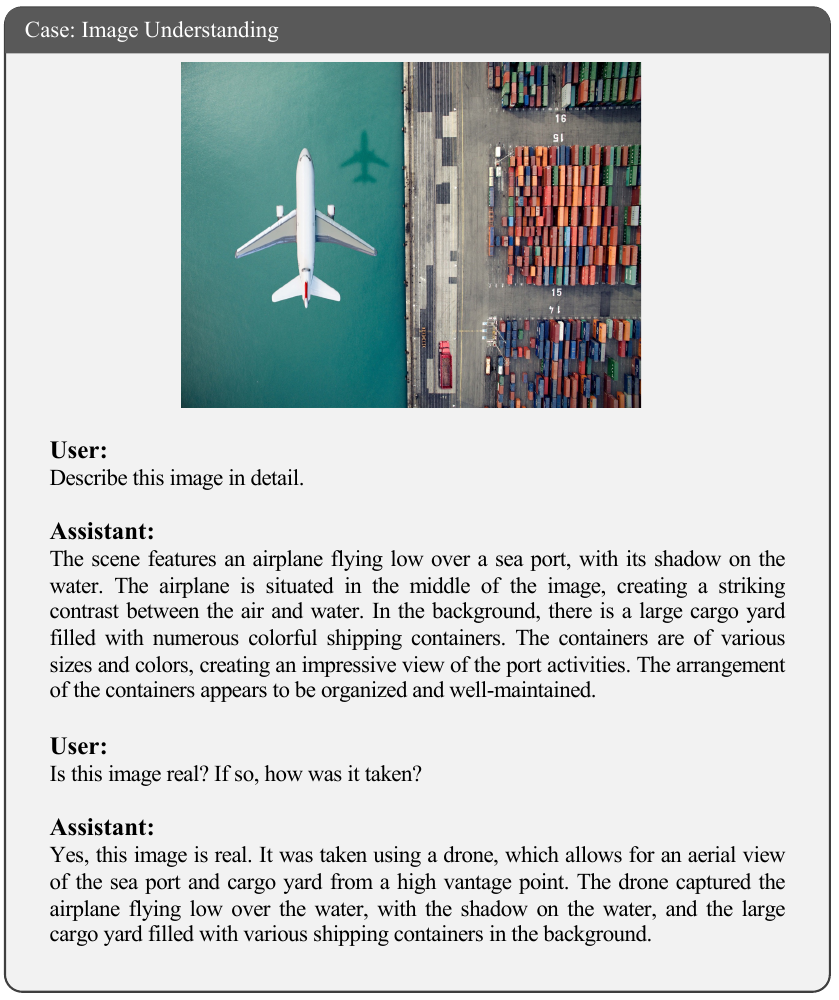}
    \caption{An example of Align-TI-2B's ability in image understanding.}
    \label{fig:case-understanding}
\end{figure*}

\begin{figure*}[htbp]
    \centering
    \includegraphics[width=0.9\linewidth]{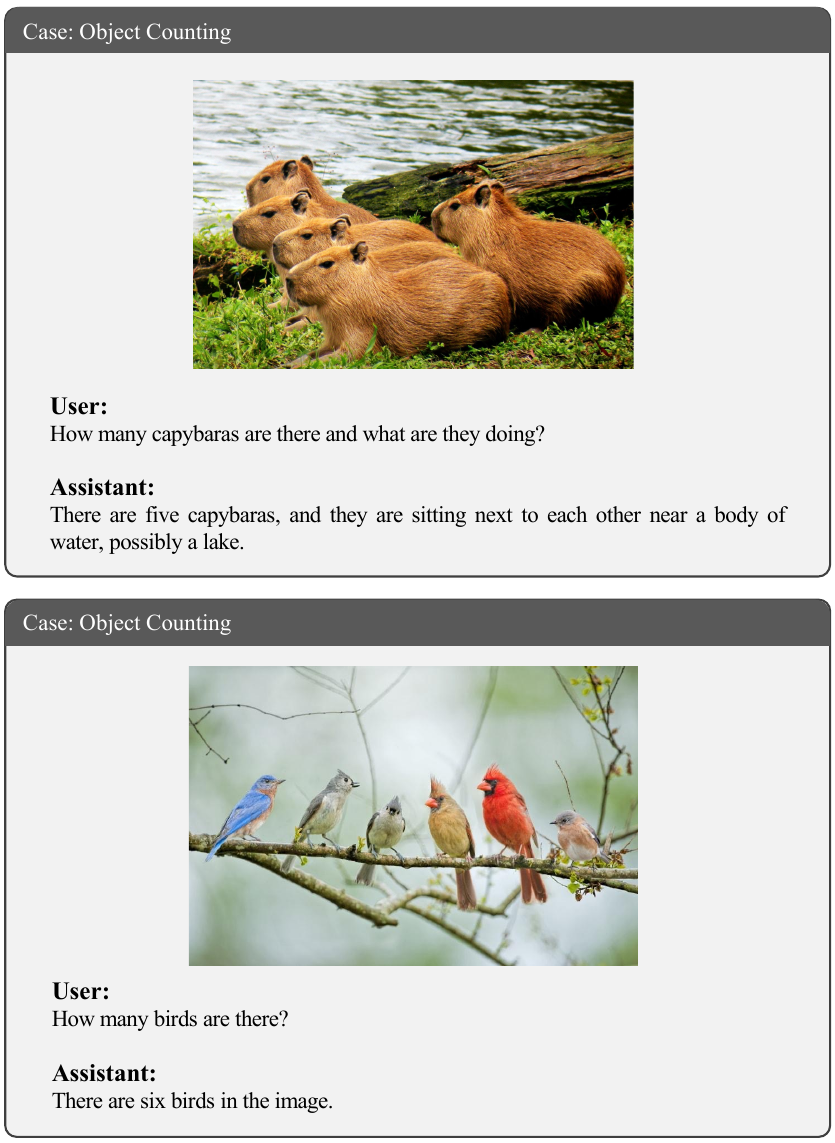}
    \caption{Examples of Align-TI-2B's capability in object counting.}
    \label{fig:case-counting}
\end{figure*}

\begin{figure*}[htbp]
    \centering
    \includegraphics[width=0.9\linewidth]{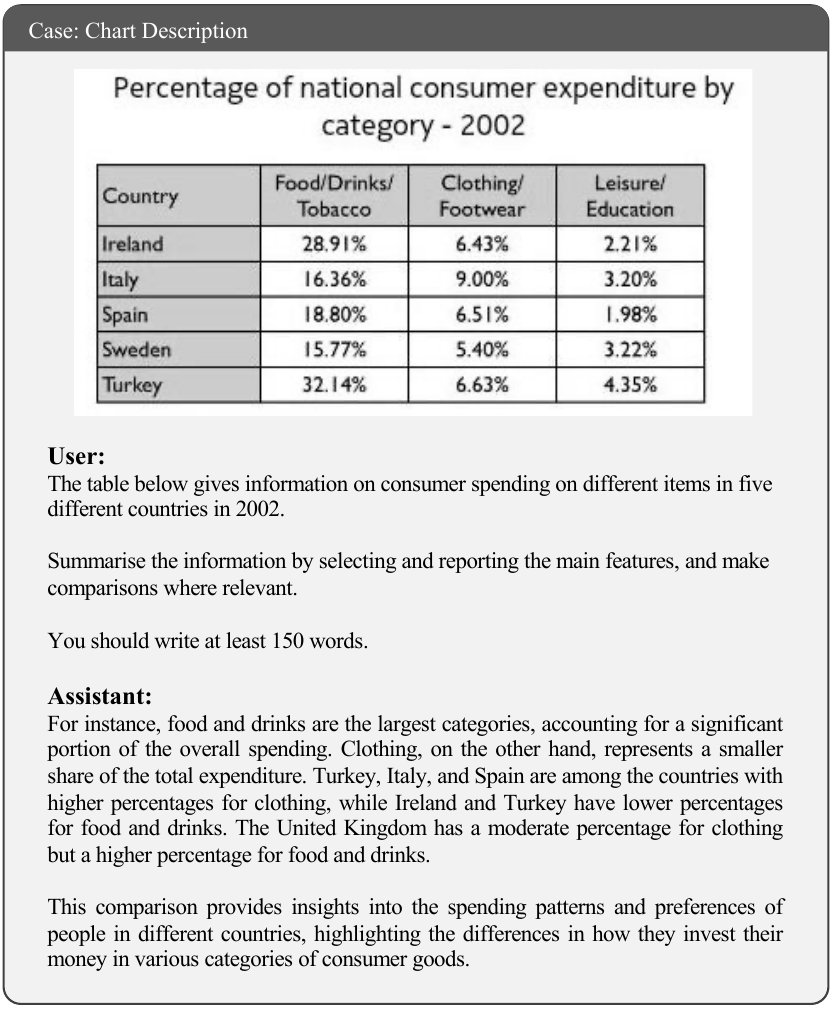}
    \caption{An example demonstrating Align-TI-2B solves chart description problem.}
    \label{fig:case-chart}
\end{figure*}

\begin{figure*}[htbp]
    \centering
    \includegraphics[width=0.9\linewidth]{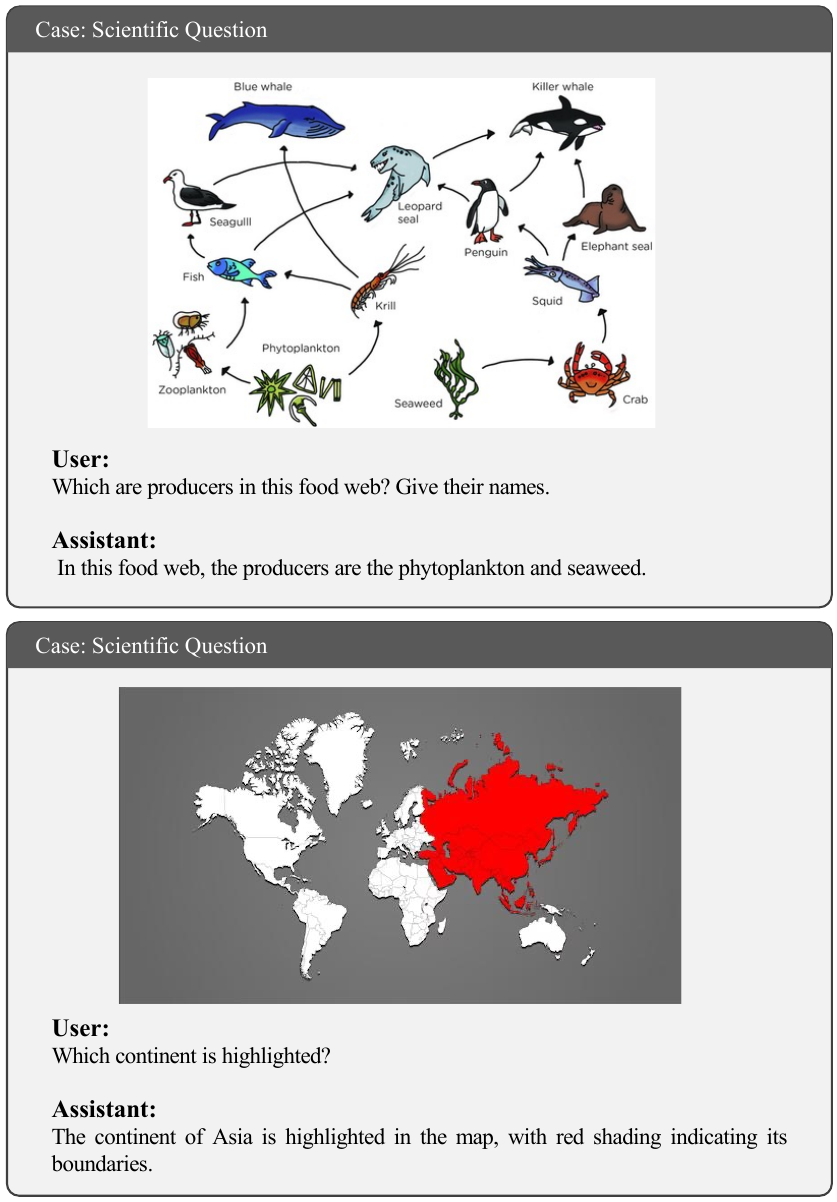}
    \caption{Examples of Align-TI-2B's capability in solving scientific questions.}
    \label{fig:case-science}
\end{figure*}

\begin{figure*}[htbp]
    \centering
    \includegraphics[width=0.9\linewidth]{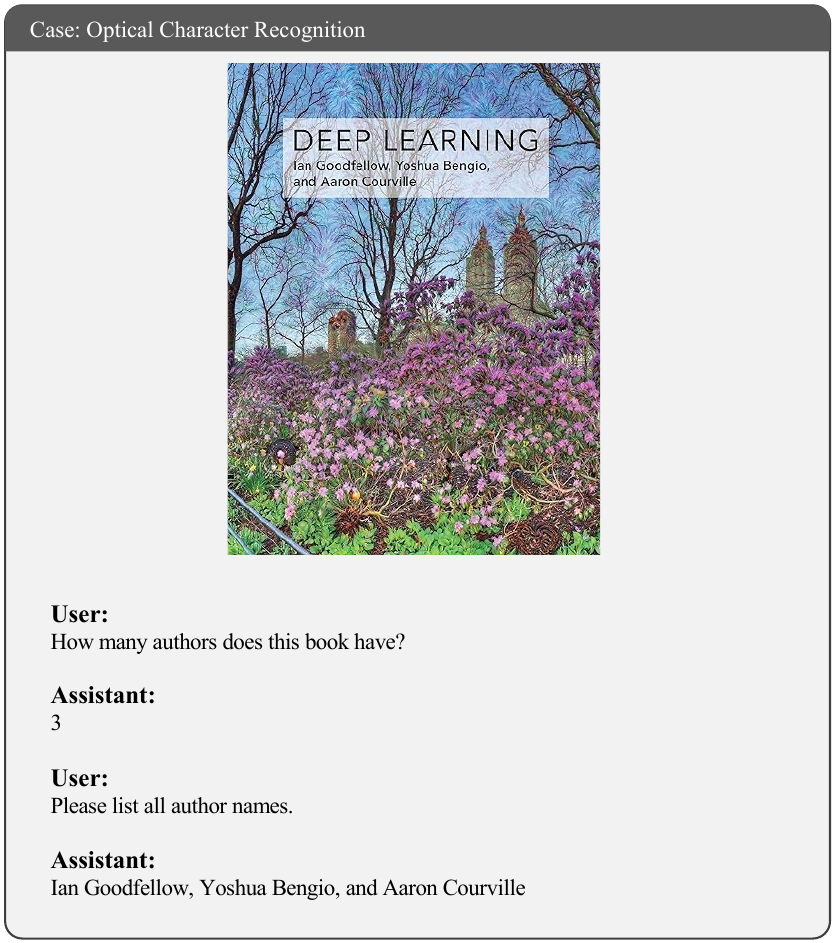}
    \caption{An example of Align-TI-2B performing Optical Character Recognition (OCR).}
    \label{fig:case-ocr}
\end{figure*}

\begin{figure*}[htbp]
    \centering
    \includegraphics[width=0.9\linewidth]{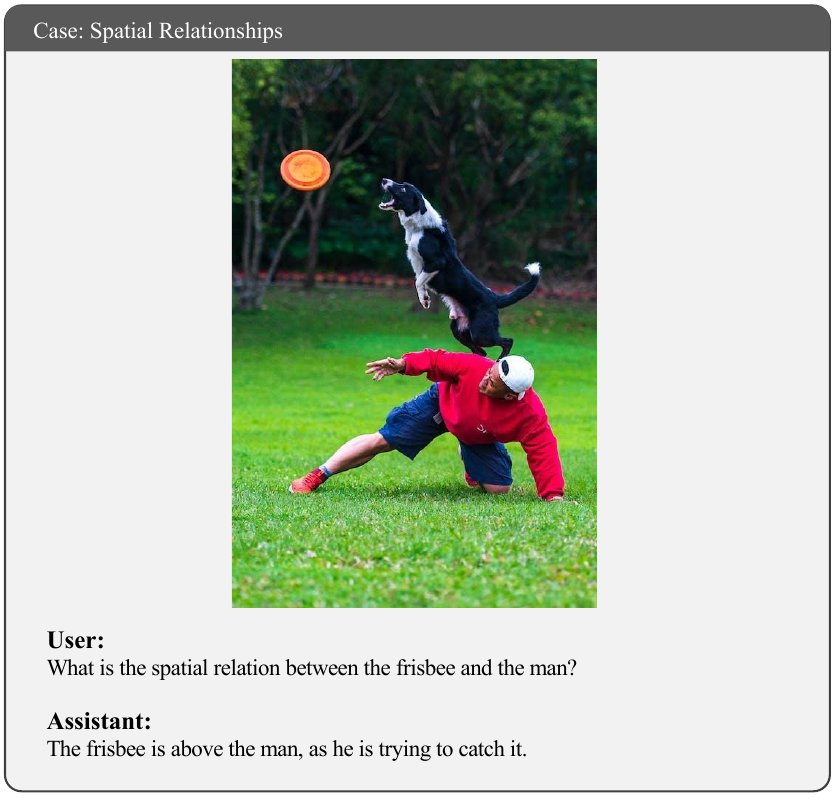}
    \caption{An example demonstrating Align-TI-2B's understanding of spatial relationships.}
    \label{fig:case-spatial}
\end{figure*}

In this section, we provide a series of case studies that qualitatively illustrate the effectiveness of our distilled Align-TI-2B model. These examples highlight the model's performance across a wide range of vision-language tasks, including image understanding (Fig.~\ref{fig:case-understanding}), object counting (Fig.~\ref{fig:case-counting}), chart description (Fig.~\ref{fig:case-chart}), solving scientific questions (Fig.~\ref{fig:case-science}), optical character recognition (Fig.~\ref{fig:case-ocr}), and spatial relationship understanding (Fig.~\ref{fig:case-spatial}).

\end{document}